\documentclass{article}

\usepackage{microtype}
\usepackage{graphicx}
\usepackage{subfigure}
\usepackage{booktabs} 
\usepackage{multirow}
\usepackage{amsmath}
\usepackage{commath}
\DeclareMathOperator{\sign}{sgn}

\usepackage{hyperref}

\usepackage[accepted]{icml2020}

\icmltitlerunning{DQI: A Guide to Benchmark Evaluation}

\begin{document}

\twocolumn[
\icmltitle{DQI: A Guide to Benchmark Evaluation}




\begin{icmlauthorlist}
\icmlauthor{Swaroop Mishra}{to}
\icmlauthor{Anjana Arunkumar}{to}
\icmlauthor{Bhavdeep Sachdeva}{to}
\icmlauthor{Chris Bryan}{to}
\icmlauthor{Chitta Baral}{to}
\end{icmlauthorlist}

\icmlaffiliation{to}{Department of Computer Science, Arizona State University}

\icmlcorrespondingauthor{Swaroop Mishra}{srmishr1@asu.edu}

\icmlkeywords{Machine Learning, ICML}

\vskip 0.3in
]



\printAffiliationsAndNotice{}  

\begin{abstract}
A `state of the art' model \textit{A} surpasses humans in a benchmark \textit{B}, but fails on similar benchmarks \textit{C}, \textit{D}, and \textit{E}. What does \textit{B} have that the other benchmarks do not? Recent research provides the answer: spurious bias. However, developing \textit{\^{A}} to solve benchmarks \textit{B} through \textit{E} does not guarantee that it will solve future benchmarks. To progress towards a model that `truly learns' an underlying task, we need to quantify the differences between successive benchmarks, as opposed to existing binary and black-box approaches. We propose a novel approach to solve this underexplored task of quantifying benchmark quality by debuting a data quality metric: DQI. 
\end{abstract}
\vspace{-2mm}
\section{Introduction}
\vspace{-1mm}
We evaluate progress in various AI domains such as NLP and Vision by building and solving increasingly harder benchmarks (and hence developing new models and architectures). Since this involves heavy investment in resources (time, money, hardware, etc.), it is reasonable to ask: \textit{Can we rely on these benchmarks?} A growing number of recent works \cite{gururangan2018annotation, schwartz2017effect, poliak2018hypothesis, kaushik2018much,le2020adversarial} reveal that models exploit spurious biases (unintended correlations between input and output \cite{torralba2011unbiased}) instead of the actual underlying features to solve many popular benchmarks. This begs a new question: \textit{How do we mitigate spurious biases in benchmarks?}

\begin{figure}[H]
\centering
    \includegraphics[height=3cm]{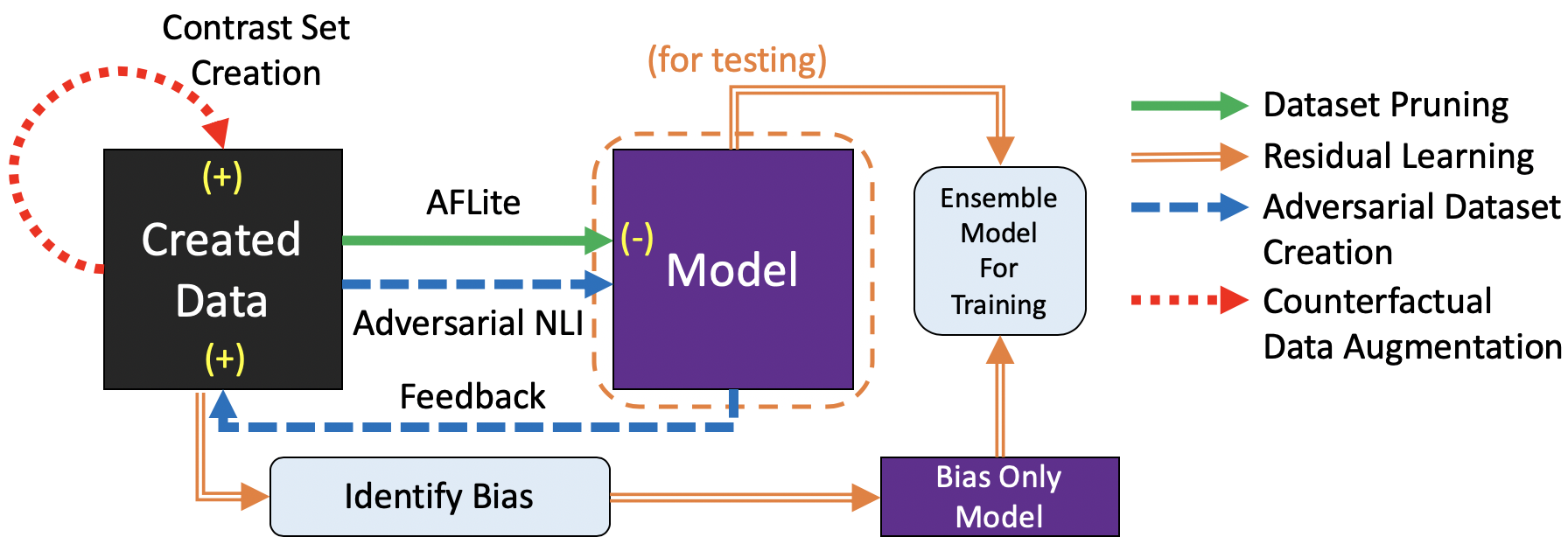}
    \vspace{-2mm}
    \caption{Existing approaches to eliminate bias}
    \label{fig:exapproaches}
    \vspace{-4mm}
\end{figure}

Recently proposed approaches that address this include dataset pruning \cite{sakaguchi2019winogrande,li2019repair,li2018resound,wang2018dataset}, residual learning \cite{clark2019don,he2019unlearn,mahabadi2019simple},  adversarial dataset creation \cite{zellers2018swag,nie2019adversarial}, and counterfactual data augmentation \cite{kaushik2019learning,gardner2020evaluating}. Each focuses on a specific part of the data-model loop, as illustrated in Figure \ref{fig:exapproaches}, but all are limited by binary evaluation: (i) accepting or rejecting a data sample created by a crowd-worker \cite{nie2019adversarial}, (ii) retaining or removing data with adversarial filtering \cite{sakaguchi2019winogrande,li2019repair,li2018resound}, (iii) augmenting only counter factual data \cite{kaushik2019learning,gardner2020evaluating}, and/or (iv) including data only if it can fool the model \cite{zellers2018swag,nie2019adversarial}. 

Binary evaluation is restrictive as it only allows inclusion or deletion of data, and further appends an overhead on human evaluators as there is uncertainty in class distinction. These approaches can also introduce new kinds of bias, and overfit to a specific model or task \cite{liu-etal-2019-inoculation}. Other limitations include: (i) wastage of resources invested in creating initial `biased' data, (ii) a dataset creator does not learn what constitutes biased data, and is likely to repeat mistakes, (iii) important aspects of bias, like its dependency on a train-test split, are ignored, (iv) model training on each iteration increases time complexity, and (v) the absence of a suitable and illustrative feedback channel. A metric \textit{quantifying benchmark quality} could address these issues, but remains underexplored. 

As a solution, we propose a novel metric: Data Quality Index (DQI), building on a recent work \cite{mishra2020dqi} which identifies potential bias parameters based on a broad survey of AI literature. We construct an empirical formula for DQI based on these parameters with seven components and a varying number of sub-components and terms (e.g., NLI has 20 sub-components and 133 terms). In our study, lower bias and higher generalizability imply higher DQI. 

DQI also captures a broad range of biases, unlike existing binary and black-box approaches (which only consider a specific set of biases). Specifically, we evaluate DQI against AFLite, a recent successful adversarial filtering approach, over NLI, QA, and RC datasets. In this paper, we focus on DQI for NLP, though our approach can be extended to other domains such as vision and speech. 

DQI is inspired by successful quality indices in domains such as power~\cite{bollen2000understanding}, air~\cite{jones1999indoor}, food~\cite{grunert2005food} and water~\cite{world1993guidelines}. On a related note, Data Shapley~\cite{ghorbani2019data} has been proposed as a metric to quantify the value of each training datum to the predictor performance, but follows a model and task-dependent approach and might fail when biases favor the predictor. So, we focus on building a generic DQI with minimal dependency on models and tasks.

\section{DQI}
DQI utilizes a generic parameter set \cite{mishra2020dqi} that captures bias properties---including origins, types and impact on performance, generalization, and robustness--- for a hierarchy of datasets ranging from NLI to Text Summarization. We abstract this set and use appropriate mathematical transformations to algorithmically compute DQI. Our intuition is simple: a data quality metric should discourage biased samples and encourage samples with higher generalization capability \cite{mishra2020our}. DQI has seven components corresponding to seven properties that cover various possible inter/intra-sample interactions in an NLP dataset, isolating those which lead to bias \footnote{More details about components and the intuition behind them are in supplemental materials}.

\paragraph{Formalization:}
Let $X$ represent a dataset with $size$ as its number of samples. $X$ has vocabulary $v$, over a set of sentences $S$, with $s$ denoting sentence lengths across $S$. Let the set of granularities across $X$ be referenced as $i \epsilon \{Words, Verbs, Adjectives, Nouns, Adverbs,$ $Bigram, Trigram, Sentences\}$, with $\nu$ representing their respective frequencies, and $c$ and $d$ hyperparameters constraining $\nu$. Let  $l$ span $S$, and $Sim_{lm}$ represent sentence similarity between the $l^{th}$ sentence and $m^{th}$ sentence of $S$, where $m$ spans $S - \{l\}$. $SIM$ is a hyperparameter that is a lower bound for $Sim_{lm}$. $e$ is a hyperparameter that depends on $size$, which is the minimum threshold for the number of sentences spanned by $m$ where $Sim_{lm}>SIM$, and $\max_{me}$ represents the similarity values for the top $e$ sentences, for every $l\epsilon S$. Let $WSim_{uv}$ stands for word similarity between the  $u^{th}$ word and the $v^{th}$ word where $u$ spans every word in a sentence $s'\epsilon S$, and $v$ spans $s'-\{u\}$ , $WSIM$ is a hyperparameter dependent on $size$ that represents the minimum threshold for $WSim_{uv}$.
Let $p$ represent sentences from one side and $h$ represent sentences from the other side, such as premise and hypothesis respectively in NLI. $ISIM$ is a hyperparameter that represents the lower bound for $Sim_{ph}$, which is the similarity between $p$ and $h$, with $s_p$ and $s_h$ representing premise and hypothesis lengths respectively. $u_w$ represents unique words in $p$ and $h$, $q$ spans the sample, and $q_p$ and $q_h$ span the premise and hypothesis respectively. Let $g$ be the upper limit for respective $i \epsilon \{Words, Verbs, Adjectives, Nouns, Adverbs,$ $Bigram, Trigram, Sentences\}$ across any indivdual label. $Count_{label}$ is a vector of size $labels$, where $labels$ represents the number of labels, which represents how many times each element of each $i$ granularity has been assigned each of the labels, $label$. Let 
$X_{train}$ and $X_{test}$ represent the train and test splits respectively of $X$.
$Sim_{train-test}$ stands for similarity between the train and test data and $SSIM$ is a hyperparameter that represents the optimal value of $Sim_{train-test}$. Let $\sign$ represent the signum function.
$DQI_C$ represents DQI components as follows:

\textbf{Vocabulary:} 

$DQI_{c1}=\frac{v(X)}{size(X)}+\sigma(s(X))*\frac{\sum_{S}\sign((s-a)(b-s))}{size(S)}$
\vspace{-4mm}
\begin{equation}
\end{equation}
\vspace{-2mm}
\textbf{Inter-Sample N-gram Frequency and Relation:} 

$DQI_{c2}=\sum_{i}(\frac{1}{\sigma (\frac{i(\nu)}{size(i)})}*\frac{\sum_{i}((\nu_{i}-c)(d-\nu_{i}))}{size(i)})$
\vspace{-4mm}
\begin{equation}
\end{equation}
\vspace{-1mm}
\textbf{Inter-Sample STS: } 

$DQI_{c3}=\frac{size(S)}{\sigma(\forall_{l} \nu_{\sign\frac{\abs{Sim_{lm}-SIM}-(Sim_{lm}-SIM)}{2}})+1}+\frac {2*size(S)}{(\sum_{l}\sum_{e}{\max\limits_{me}\frac{}{}(\abs{Sim_{lm}-SIM}}-(Sim_{lm}-SIM)))+1}$
\vspace{-4mm}
\begin{equation}
\end{equation}
\vspace{-1mm}
\textbf{Intra-Sample Word Similarity: } 

$DQI_{c4}=\frac{size(S)}{\sum_{S}({\forall_{l}\abs{\frac{\sum_{m}WSim_{uv}}{length(s')}-WSIM}})+1}$
\vspace{-4mm}
\begin{equation}
\end{equation}
\vspace{-1mm}
\textbf{Intra-Sample STS: } 

$DQI_{c5}=\frac{size(X)}{\sum_{X}\left | \forall_{p}\forall_{h}Sim_{ph}-ISIM \right |+1}
+\frac{size(X)}{\sum_{X}\left | (s_p-s_h) \right |+1}
+\frac{\sigma \left ( \left | s_{p}-s_{h} \right | \right )}{size(X)}
+\frac{\sigma(\forall_{p} \forall_{h} Sim_{ph} )}{size(X)}
+\frac{\sum_X(\frac{s_p+s_h}{\forall_{uw}\sum_{q} \sign(2-\nu_{q})})}{size(X)}
+\frac{\sum_X(\frac{1}{\forall_{uw}\sum_{u\epsilon q_h} \max\limits_{v \epsilon q_p}WSim_{uv}})}{size(X)}
$
\vspace{-4mm}
\begin{equation}
\end{equation}
\vspace{-1mm}
\textbf{N-Gram Frequency per Label: }

$DQI_{c6}=\sum_{labels}(\sum_{i}\frac{1}{\sigma(\frac{i(\nu)}{size(i)})}*\frac{\sum_{i}((g-\nu_{i}))}{size(i)}+\frac{size(X_{label})}{(\sum_{X_{label}}(\left|(s_p-s_h)\right|))+1}+\frac{\sigma(\left|(s_p-s_h)\right|)}{size(X_{label})})+\sum_{i}\frac{size(i(X))}{(\sum_{i(X)}\sigma(\forall_{X}\frac{(\left|{1-Count_{label}}\right|-(1-Count_{label}))}{2}))+1}$
\vspace{-4mm}
\begin{equation}
\end{equation}
\vspace{-1mm}
\textbf{Inter-Split STS: }

$DQI_{c7}=\frac{size(X_{test})}{(\sum_{test}{\left|\max\limits_{X_{train}}{Sim_{train-test}}-SSIM\right|})+1}$
\vspace{-4mm}
\begin{equation}
\end{equation}
\vspace{-1mm}
We propose the empirical formula of DQI as a function of all components.

$DQI=f(DQI_1, DQI_2, DQI_3, DQI_4, DQI_5,DQI_6,\\DQI_7)$
\vspace{-6mm}
\begin{equation}
\end{equation}

Since $f$ depends on both task and dataset, it needs to be experimentally tuned.

\section{Comparing Performance Against AFLite} \label{dqieval}
We apply DQI to compare its performance to that of AFLite on four datasets: SNLI~\cite{bowman2015large}, MNLI~\cite{williams2017broad}, SQUAD 2.0~\cite{rajpurkar2018know}, and Story CLOZE Task~\cite{mostafazadeh2016corpus}. AFLite divides samples into \textit{good} and \textit{bad} splits, i.e. samples retained and removed on filtering. Mishra et. al. \cite{mishra2020dqi} show that SNLI contains a large number of  artifacts, and that the Story CLOZE Task also has a significant number of artifacts.  MNLI and SQUAD 2.0 are shown to have a relatively smaller number of artifacts, thus ensuring an adversarial evaluation of DQI. We tune hyperparameters on 0.01\% of data manually in a supervised manner, mimicking how humans learn quickly from a few samples.\footnote{Detailed tuning results with various hyperparameters are in supplemental materials.}
We perform two types of evaluation: (i) overall analysis of 133 terms, and 7 components to ascertain AFLite intricacies, and (ii) individual sample analysis across the most sensitive sub-components. 

\subsection{Overall Analysis: }
By applying DQI to AFLite\footnote{Detailed analysis of each DQI sub-component and experimental results for all datasets are in Supplemental Materials.}, we can analyze where AFLite fails and succeeds at sample splitting.

\textbf{AFLite Failures:} We specifically examine language features that AFLite fails to appropriately consider as artifacts. The DQI formulas are constructed such that the \textit{good} split is expected to have higher sub-component values than the \textit{bad} split.

\textbf{Sentence length:} We expect variation of sentence lengths to be high, as length has been found to be an important parameter related to bias in SNLI \cite{mishra2020dqi}. We find that even though the second and third sub-components of the \textit{Vocabulary} component are higher for the good split, the difference is less than expected. Sentence length variation follows a similar pattern for each split. This is confirmed by calculating the percentage differences of sentence lengths between the splits. The takeaway is that AFLite likely does not appropriately consider data with sentence length associated bias, as we would otherwise expect to see sentences with outlier length values mainly placed in the \textit{bad} split. This is further supported by sub-component three (fails for neutral and contradiction labels) and sub-component four (fails for contradiction label) of the \textit{N-gram Frequency per Label} component---responsible for ensuring that models do not overfit towards a fixed-length difference.

\textbf{Sentence Similarity:} For the \textit{Inter-sample STS} component, sub-component one dictates that the number of sentences that cross the threshold set for spurious bias should have lower variance: if the distributions of similarity for all sentences are skewed, this leads to spurious bias. We find that the \textit{bad} split outperforms the \textit{good} split, which indicates that AFLite might not be not considering imbalance due to sentence similarity. 

\textbf{Premise-Hypothesis Similarity}
The \textit{Intra-sample STS} component quantifies: (i)  how far premise- hypothesis pairs are from a particular similarity threshold, (ii) how much the length variation, word overlap, and maximum word similarity between premise and hypothesis are, and (iii) how much is the variation in similarities across all pairs in the dataset. We expect significant\footnote{Significance is defined as values of order greater than e-03 for this component.} differences for sub-components between the \textit{good} and \textit{bad} splits. However, both sub-component and overall component values do not show a significant difference across splits. This is surprising, as this component captures several major bias-related parameters \cite{mishra2020dqi}. This indicates AFLite might not be accurately filtering data with high premise-hypothesis similarity and length difference.

\textbf{Bigrams, Trigrams:} 
We expect a non-skewed distribution of granularities both within and across labels. We find that the first sub-component for \textit{N-gram Frequency per Label} fails for bigrams, and trigrams. AFLite is likely not handling these granularities appropriately. For bigrams and trigrams, the fifth sub-component again has a lower value for the \textit{good} split, indicating AFLite is not effectively identifying artifacts for bigrams and trigrams.

\textbf{Neutral Category:}
For the \textit{N-gram Frequency per Label} component, the second sub-component fails in the neutral label for the sentence, adjective, adverb, verb, bigram, and trigram granularities. This indicates that AFLite is potentially not filtering appropriately for neutral category samples.

\textbf{Train-Test Split:}
For the \textit{Inter-Split STS} component,we find no significant difference in train-test similarity between the \textit{good} and \textit{bad} splits, though it is expected that the \textit{bad} split will show much higher similarity, as inter-split similarity has been identified as an important source of bias in SNLI \cite{mishra2020dqi}. This indicates AFLite is  poten-

\begin{figure*}
    \centering
    \includegraphics[width=2\columnwidth]{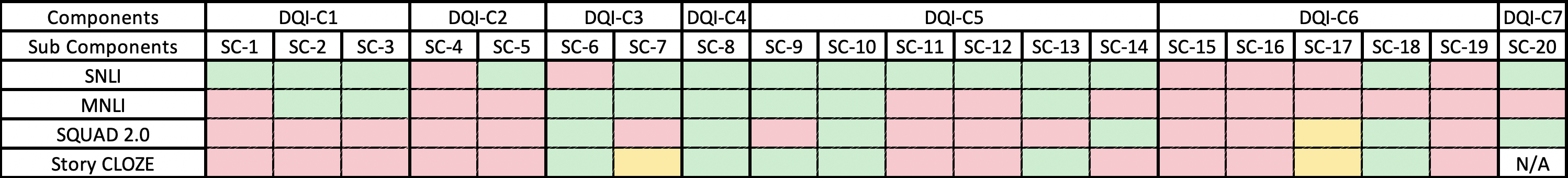}
    \caption{Summarized results for SNLI, MNLI, SQUAD 2.0, and Story CLOZE Task. Green indicates that the sub-component, \textit{SC}, has a higher value for the \textit{good} split, and red for the \textit{bad} split. Yellow indicates that a tie is seen between the \textit{good} and \textit{bad} splits. Inter-Split Similarity is not evaluated in Story CLOZE Task due to the  absence of training data. }
    \label{fig:otherdata}
\end{figure*}

tially not properly incorporating artifacts related to the train-test split, such as data leakage.

\textbf{AFLite Pass Cases:}
For the \textit{Vocabulary} component, the \textit{good} split has a higher overall value than the \textit{bad} split. Of the three sub-components in this component, the first shows the most significant difference. The granularity variation in the \textit{Inter-Sample N-Gram Frequency and Relation} component passes for all granularities except sentences, which we attribute to lower repetition of sentences compared to the other granularities. We also calculate this sub-component without normalization and find that it holds for sentences without normalization; the second sub-component passes in all cases. The second sub-component for \textit{Inter-Sample STS} also passes. We also observe that the \textit{Intra-Sample Word Similarity} component passes, indicating that AFLite captures \textit{Word Noise} in SNLI. We note that contradiction samples seem more prone to spurious bias, due to a high ratio between the \textit{bad} and \textit{good} split sample counts in comparison to the entailment and neutral labels.

\textbf{Other Datasets:}
Figure \ref{fig:otherdata} summarizes results for SNLI, MNLI, SQUAD 2.0, and Story CLOZE Task.\footnote{Detailed results are in supplemental materials} The number of sub-components for which the \textit{good} split has higher DQI values than the \textit{bad} split reduces as we move in order between SNLI, Story CLOZE Task, MNLI, and SQUAD 2.0. This is likely due to the decrease in the number of artifacts.
\begin{figure}[h]
    \centering
    \includegraphics[height=4.5cm]{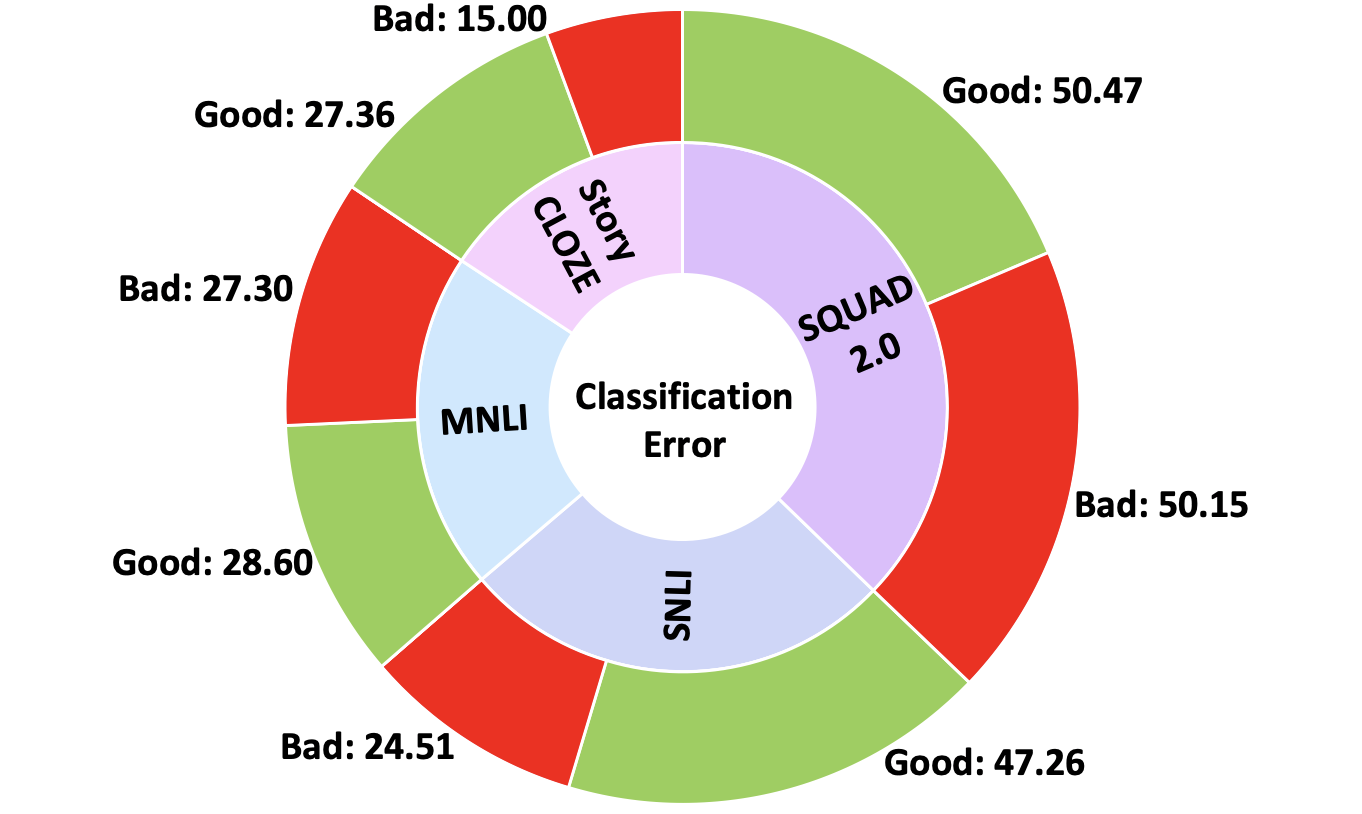}
    \caption{Misclassification percentages of AFLite, post evaluation using word overlap, word similarity and sentence length.}
    \label{fig:percent}
\end{figure}

\textbf{3.2. Sample-Wise Analysis}

We individually evaluate a subset of samples to quantify inconsistencies in AFLite. We set a minimum threshold value for DQI components to bin samples in the \textit{good} split, by following the same steps as that of hyperparameter tuning (mentioned at the top of this section). Next, we calculate the DQI of samples in the \textit{good} and \textit{bad} splits and look for inconsistencies. Figure~\ref{fig:percent} summarizes the results, showing that 47.26\% and 24.51\% of SNLI samples are misclassified in the \textit{good} and \textit{bad} splits. The percentages for the other datasets are MNLI 28.60\%/27.30\%, SQUAD 2.0 50.47\%/50.15\%, and Story CLOZE Task 27.36\%/15.00\%. 
\section{Discussion: Towards a Paradigm Shift in Benchmarks and Models}
DQI's ability to quantify data quality can: (i) be leveraged to repair biased legacy datasets, (ii) provide realtime feedback to crowdworkers when creating samples for benchmarks, (iii) provide flexibility in controlling the `hardness' of a benchmark by tuning relevant sub-components out of the 133 terms, (iv) help better utilize the investment of resources in creating datasets, as it does not require the deletion of biased data at a later stage, and (v) help understand which language properties are important to solve a dataset.


\section{Conclusion} 
We introduce a novel metric Data Quality Index (DQI) to evaluate the quality of data in benchmarks. We build upon existing studies on bias and propose a formula for generic DQI. In contrast to existing binary and black-box approaches that only cover a specific set of biases, DQI captures a broad range of biases. DQI can serve as an automated mechanism to provide continuous feedback, neither overloading humans nor risking the possibility of bias associated with human validation. We use DQI to evaluate AFLite, a state of the art approach for adversarial filtering of NLP benchmarks. Our results show that DQI captures varieties of biases that AFLite does not capture. We show the efficacy of DQI in datasets spanning NLI, QA, and RC tasks. DQI already empowers the novel benchmarking paradigms in a series of recent works, and can further serve to inspire and validate the next generation of datasets and models.
\section*{Acknowledgements}

We thank the anonymous reviewers for their thoughtful feedback. We also thank Jason Yalim and ASU HPC for their consistent support. The support of DARPA SAIL-ON program (W911NF2020006)  is gratefully acknowledged.
\bibliography{neurips_2020.bib}
\bibliographystyle{abbrvnat}
\vspace{-3mm}

\section{Supplementary}

\begin{figure}[H]
    \centering
    \includegraphics[width=\columnwidth]{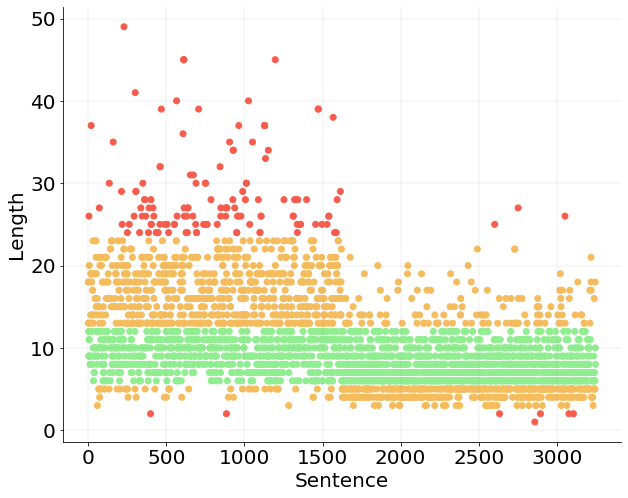}
    \includegraphics[width=\columnwidth]{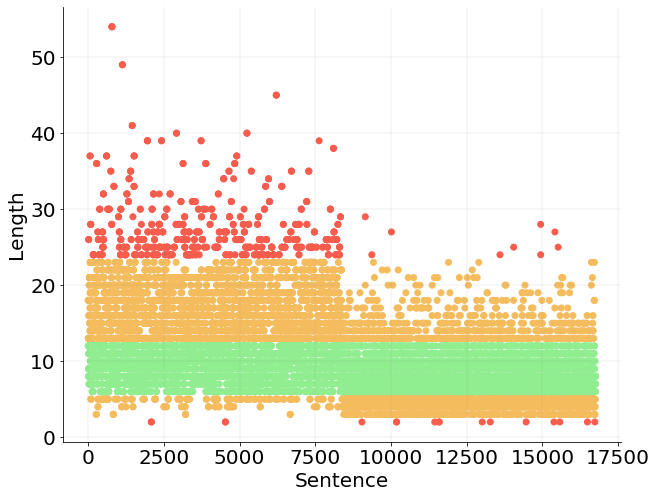}
    \vspace{-4mm}
    \caption{Sentences in SNLI visualized according to whether AFLite puts it in the \textit{good} or \textit{bad} split respectively. Each sentence is one dot; its vertical position denotes its length, and color indicates its DQI rating based on its Vocabulary component (green = good, orange = acceptable, red = bad).}
\vspace{-8mm}
\end{figure}

\begin{figure}[H]
    \centering
    \includegraphics[width=5cm,height=4cm]{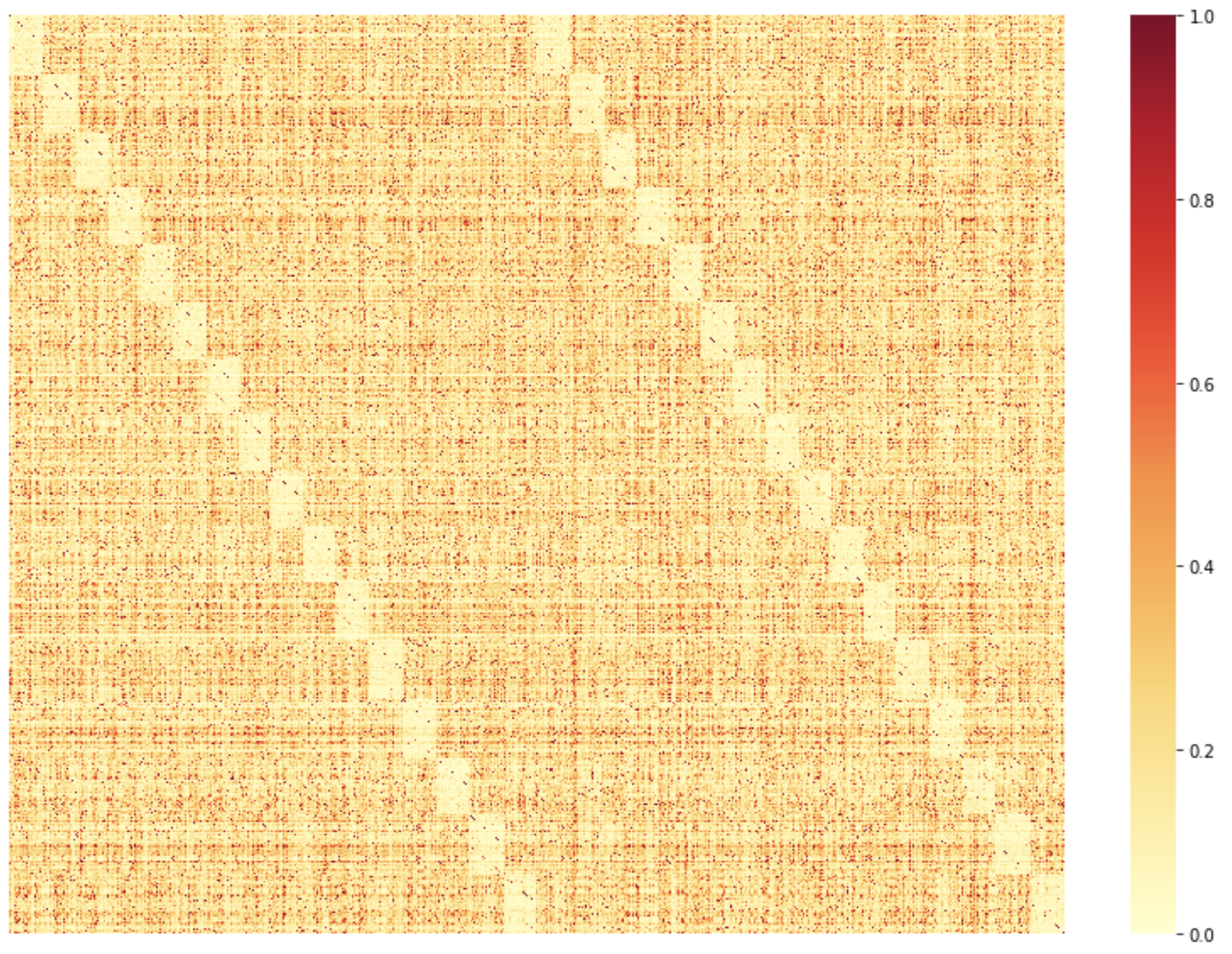}
    \includegraphics[width=5cm,height=4cm]{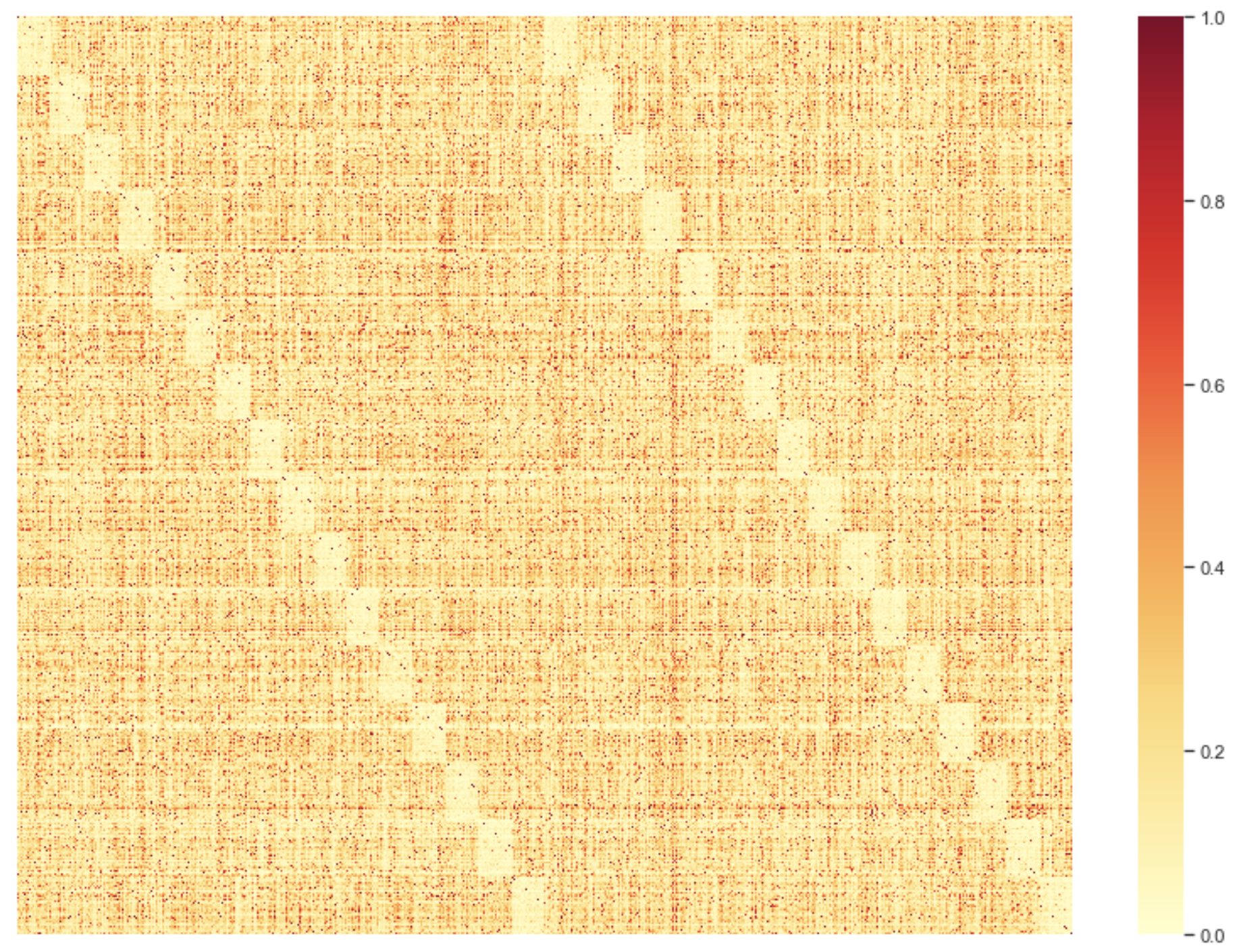}
    \includegraphics[width=5cm,height=4cm]{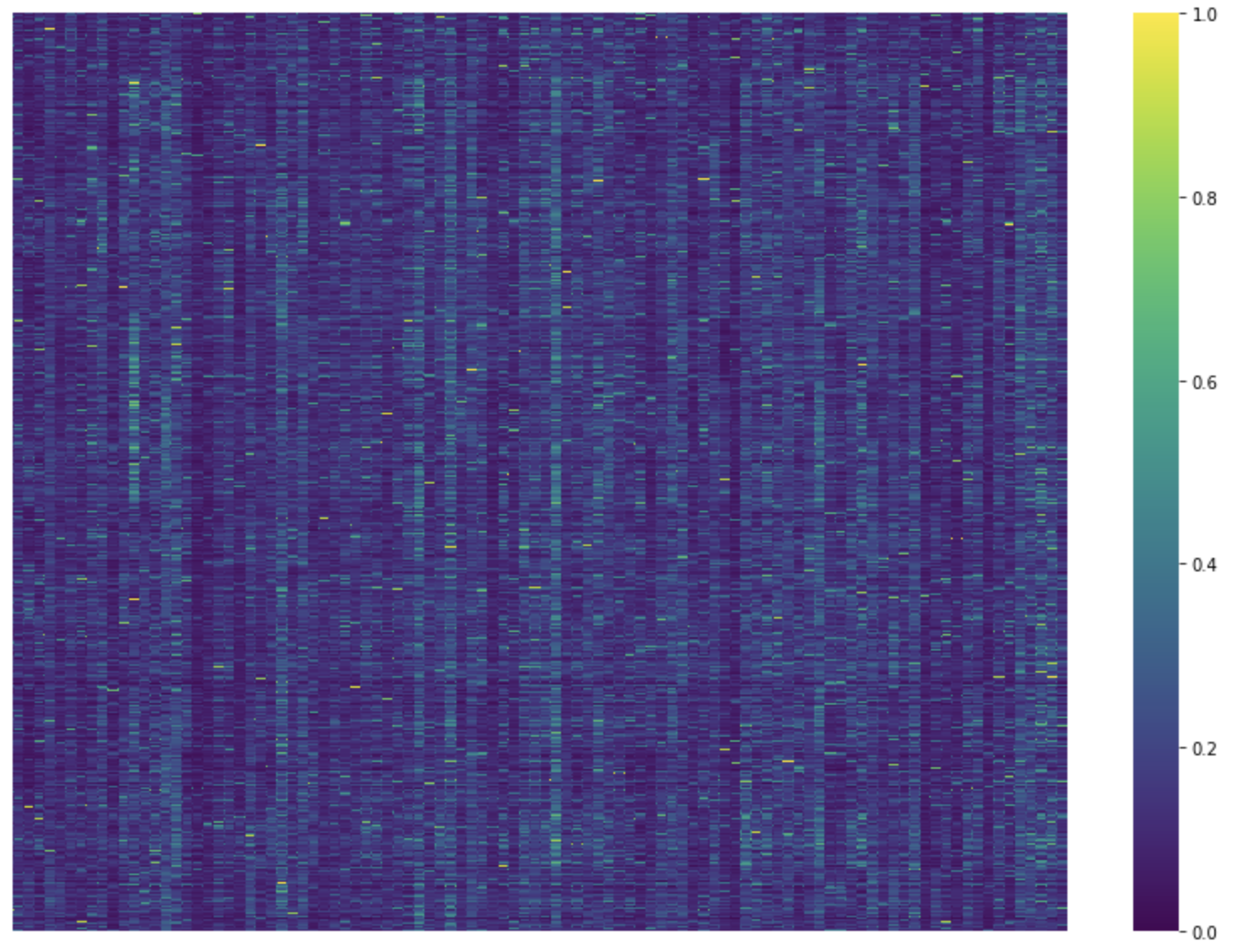}
    \includegraphics[width=5cm,height=4cm]{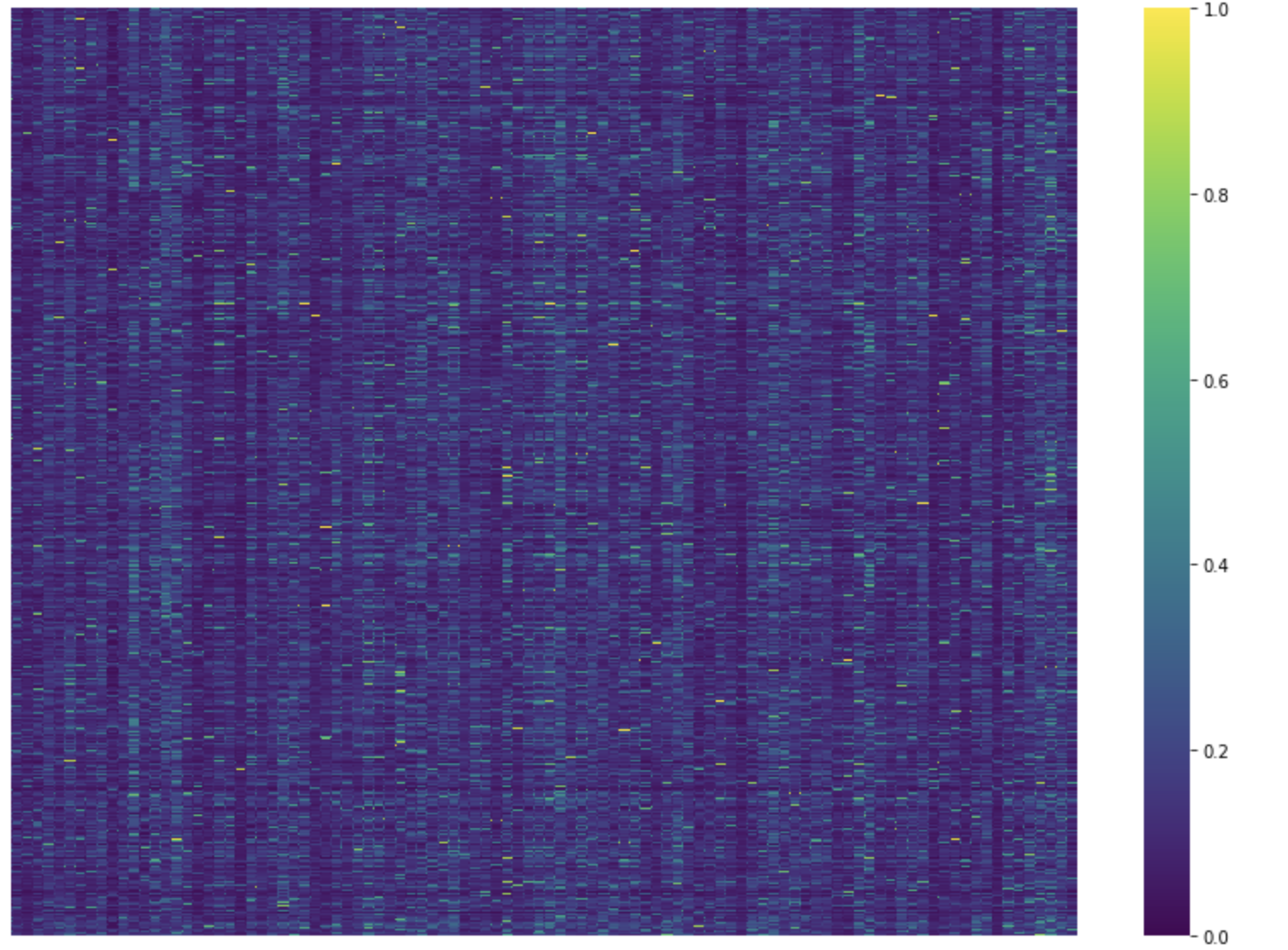}
    \caption{Semantic Textual Similarity plots where both row and column span all sentences in the dataset for C3 and rows represent train split and columns represent test split for C7. Color represents the similarity value. For C3 in the top two figures for the \textit{good} and \textit{bad} splits respectively, yellow represents zero similarity, and as the color moves towards red, the similarity increases. For C7 in the bottom two figures for the \textit{good} and \textit{bad} splits respectively, blue represents zero similarity, and as the color moves towards yellow, the similarity increases.}
\vspace{-6mm}
\end{figure}

\textbf{Vocabulary:}

\begin{table}[H]
\scriptsize
\centering
\begin{tabular}{lllll}
\hline
\textbf{Term} & \textbf{T1} & \textbf{T2} & \textbf{T3} & \textbf{DQI C1} \\
\hline
\textbf{Good} & \textbf{1.8996} & \textbf{6.0409}   & \textbf{0.9532} & \textbf{7.6578} \\
\textbf{Bad}  & 0.6416 & 5.8135   & 0.9494 & 6.1609 \\\hline
              &            &            &            &                
\end{tabular}
\vspace{-4mm}
\caption{SNLI Sub-Component and Overall Values for $DQI_{c1}$}.
\end{table}

\begin{table}[H]
\scriptsize
\centering
\begin{tabular}{lllll}
\hline
\textbf{Term} & \textbf{T1} & \textbf{T2} & \textbf{T3} & \textbf{DQI C1} \\
\hline
\textbf{Good} & 1.6177 & \textbf{104.6542}   & \textbf{0.7550} & \textbf{80.6316} \\
\textbf{Bad}  & \textbf{7.4100} & 14.1068   & 0.6020 & 15.9023 \\\hline
              &            &            &            &                
\end{tabular}
\vspace{-4mm}
\caption{MNLI Sub-Component and Overall Values for $DQI_{c1}$}.

\begin{tabular}{lllll}
\hline
\textbf{Term} & \textbf{T1} & \textbf{T2} & \textbf{T3} & \textbf{DQI C1} \\
\hline
\textbf{Good} & 1.7715 & 71.3947   & -0.0023 & 1.6073 \\
\textbf{Bad}  & \textbf{11.1550} & \textbf{73.3092}   & \textbf{-0.001} & \textbf{11.1476} \\\hline
              &            &            &            &                
\end{tabular}
\vspace{-4mm}
\caption{SQUAD 2.0 Sub-Component and Overall Values for $DQI_{c1}$}.

\begin{tabular}{lllll}
\hline
\textbf{Term} & \textbf{T1} & \textbf{T2} & \textbf{T3} & \textbf{DQI C1} \\
\hline
\textbf{Good} & 3.3010 & 13.4569  & 0.2772 & 7.0313 \\
\textbf{Bad}  & \textbf{4.7675} & \textbf{13.4895}   & \textbf{0.2839} & \textbf{8.5972} \\\hline
              &            &            &            &                
\end{tabular}
\vspace{-4mm}
\caption{Story-CLOZE Sub-Component and Overall Values for $DQI_{c1}$}.
\vspace{-8mm}
\end{table}

\textbf{Inter-Sample N-Gram Frequency and Relation:}

\begin{table} [H]
\centering
\scriptsize
\begin{tabular}{lllll}
\hline
\textbf{Granularity} & \textbf{Split} & \textbf{T1} & \textbf{T2} & \textbf{Contribution} \\
\hline
\textbf{Words} &\textbf{Good}& \textbf{121.9512} & \textbf{0.7269}   & \textbf{88.6463} \\
 &\textbf{Bad}& 52.3560 & 0.6500   & 34.0314 \\\hline
\textbf{Adjectives} &\textbf{Good}& \textbf{31.7460} & \textbf{0.2966}   & \textbf{9.4159} \\
 &\textbf{Bad}& 16.9205 & 0.3590  & 6.0745 \\\hline
\textbf{Adverbs} &\textbf{Good}& \textbf{21.0970} & \textbf{0.1847}  & \textbf{3.8966} \\
 &\textbf{Bad}& 10.7875 & 0.1732   & 1.8684 \\\hline
\textbf{Verbs} &\textbf{Good}& \textbf{43.6681} & \textbf{0.2349} & \textbf{10.2576} \\
 &\textbf{Bad}& 16.5289 & 0.1893  & 3.1289 \\\hline
\textbf{Nouns}&\textbf{Good}& \textbf{49.2611} & \textbf{0.4351}  & \textbf{21.4335} \\
 &\textbf{Bad}& 21.0084 & 0.3685   & 7.7416 \\\hline
\textbf{Bigrams} &\textbf{Good}& \textbf{1296.3443} & \textbf{0.9374}  & \textbf{1215.1931} \\
 &\textbf{Bad}& 873.2862 & 0.9355   & 816.9592 \\\hline
\textbf{Trigrams} &\textbf{Good}& \textbf{7686.3951} & \textbf{0.9546} & \textbf{7337.4328} \\
  &\textbf{Bad}& 6119.9510 & 0.9422  & 5766.2178 \\\hline
\textbf{Sentences} &\textbf{Good}& 9070.7819 & \textbf{0.6607} & \textbf{5993.0656} \\
 &\textbf{Bad}& \textbf{14537.0541} & 0.2705   & 3932.2731\\\hline
 \textbf{Sentences} &\textbf{Good}& \textbf{3.0656} & \textbf{0.6607} & \textbf{3.7263} \\
 \textbf{(Not Normalized)} &\textbf{Bad}& 1.2655 & 0.2705   & 1.0607\\\hline
\textbf{DQIC2} &\textbf{Good}& - & - & \textbf{8668.3012} \\
  &\textbf{Bad}& -& - & 6636.3641 \\\hline
  
              &            &            &        &                   
\end{tabular}
\vspace{-4mm}
\caption{SNLI Sub-Component and Overall Values for $DQI_{c2}$}
\vspace{1mm}

\begin{tabular}{lllll}
\hline
\textbf{Granularity} & \textbf{Split} & \textbf{T1} & \textbf{T2} & \textbf{Contribution} \\
\hline
\textbf{Words} &\textbf{Good}& 299.2489 & 0.9223   & 275.9972 \\
 &\textbf{Bad}& \textbf{1026.2828} & \textbf{1.0000}   & \textbf{1026.2828} \\\hline
\textbf{Adjectives} &\textbf{Good}& 147.7382 & \textbf{1.0000}   & 147.7382 \\
 &\textbf{Bad}& \textbf{333.8001} & \textbf{1.0000}  & \textbf{333.8001} \\\hline
\textbf{Adverbs} &\textbf{Good}& 14.9467 & 0.5166  & 7.7214 \\
 &\textbf{Bad}& \textbf{54.2488} & \textbf{0.7318}   & \textbf{39.6992} \\\hline
\textbf{Verbs} &\textbf{Good}& 76.0906 & 0.6893 & 52.4492 \\
 &\textbf{Bad}& \textbf{182.7695} & \textbf{0.7130}  & \textbf{130.3146} \\\hline
\textbf{Nouns}&\textbf{Good}& 225.1162 & \textbf{0.9726}  & 218.9480 \\
 &\textbf{Bad}& \textbf{477.5051} & 0.9704   & \textbf{463.3709} \\\hline
\textbf{Bigrams} &\textbf{Good}& 4394.8945 & \textbf{1.0000}  & 4394.8945 \\
 &\textbf{Bad}& \textbf{5615.4581} & \textbf{1.0000}   & \textbf{5615.4581} \\\hline
\textbf{Trigrams} &\textbf{Good}& 16628.8816 & 0.9907 & 16474.2330 \\
  &\textbf{Bad}& \textbf{35285.2261} & \textbf{0.9735}  & \textbf{34350.1676} \\\hline
\textbf{Sentences} &\textbf{Good}& \textbf{15197.5684} & 0.0049 & 74.4680 \\
 &\textbf{Bad}& 11085.6756 & \textbf{0.9680}   & \textbf{10730.9339}\\\hline
 \textbf{Sentences} &\textbf{Good}& 1.2314 & 0.0049 & 0.0060 \\
 \textbf{(Not Normalized)} &\textbf{Bad}& \textbf{11.1732} & \textbf{0.9680}   & \textbf{10.8156}\\\hline
\textbf{DQIC2} &\textbf{Good}& - & - & 21646.4558 \\
  &\textbf{Bad}& -& - & \textbf{52700.84312} \\\hline
  
              &            &            &        &                   
\end{tabular}
\vspace{-4mm}
\caption{MNLI Sub-Component and Overall Values for $DQI_{c2}$}
\vspace{-4mm}
\end{table}

\paragraph{ }
\paragraph{ }
\paragraph{ }
\paragraph{ }

\begin{table}[H]
\scriptsize
\centering
\begin{tabular}{lllll}
\hline
\textbf{Granularity} & \textbf{Split} & \textbf{T1} & \textbf{T2} & \textbf{Contribution} \\
\hline
\textbf{Words} &\textbf{Good}& 138.6878 & \textbf{0.6744}   &  93.5310 \\
 &\textbf{Bad}& \textbf{615.0626} & 0.6224   & \textbf{382.8149} \\\hline
\textbf{Adjectives} &\textbf{Good}& 37.0775 & \textbf{1.0000}   & 37.0775 \\
 &\textbf{Bad}& \textbf{161.0191} & \textbf{1.0000 } & \textbf{161.0191} \\\hline
\textbf{Adverbs} &\textbf{Good}& 4.0080 & 0.7473  & 2.9951 \\
 &\textbf{Bad}& \textbf{18.7378} & \textbf{0.7610}   & \textbf{14.2594} \\\hline
\textbf{Verbs} &\textbf{Good}& 30.1469 & 0.9051 & 27.2859 \\
 &\textbf{Bad}& \textbf{152.9500} & \textbf{0.9372}  & \textbf{143.3447} \\\hline
\textbf{Nouns}&\textbf{Good}& 58.5576 & \textbf{1.0000}  & 58.5576 \\
 &\textbf{Bad}& \textbf{255.8677} & 1.0000   & \textbf{255.8677} \\\hline
\textbf{Bigrams} &\textbf{Good}& 1665.8142 & \textbf{0.9763}  & 1626.3344 \\
 &\textbf{Bad}& \textbf{4563.8191} & 0.9755   & \textbf{4452.0055} \\\hline
\textbf{Trigrams} &\textbf{Good}& 20526.6346 & \textbf{1.0000} & 20526.6346 \\
  &\textbf{Bad}& \textbf{39155.8925} & 0.9821  & \textbf{38455.0020} \\\hline
\textbf{Sentences} &\textbf{Good}& \textbf{4811.1347} & -0.0013 & -6.2544 \\
 &\textbf{Bad}& 1996.9248 & \textbf{0.2460}   & \textbf{491.2435}\\\hline
 \textbf{Sentences} &\textbf{Good}& 0.3991 & -0.0013 & -0.0005 \\
 \textbf{(Not Normalized)} &\textbf{Bad}& \textbf{1.3043} & \textbf{0.2460}   & \textbf{0.3208}\\\hline
\textbf{DQIC2} &\textbf{Good}& - & - & 22366.1613 \\
  &\textbf{Bad}& -& - & \textbf{44355.87788} \\\hline
  
              &            &            &        &                   
\end{tabular}
\vspace{-4mm}
\caption{SQUAD 2.0 Sub-Component and Overall Values for $DQI_{c2}$}
\vspace{1mm}

\begin{tabular}{lllll}
\hline
\textbf{Granularity} & \textbf{Split} & \textbf{T1} & \textbf{T2} & \textbf{Contribution} \\
\hline
\textbf{Words} &\textbf{Good}& \textbf{396.9190} & \textbf{0.3661}   & \textbf{145.3120} \\
 &\textbf{Bad}& 52.3560 & 0.3239   & 16.9581 \\\hline
\textbf{Adjectives} &\textbf{Good}& \textbf{77.3987} & \textbf{0.8307}   & \textbf{64.2951} \\
 &\textbf{Bad}& 70.2610 & 0.8020  & 56.3493 \\\hline
\textbf{Adverbs} &\textbf{Good}& 17.3230 & 0.4292  & 7.4350 \\
 &\textbf{Bad}& \textbf{27.8482} & \textbf{0.6178}   & \textbf{17.2046} \\\hline
\textbf{Verbs} &\textbf{Good}& 59.4638 & \textbf{0.5936} & \textbf{35.2977} \\
 &\textbf{Bad}& \textbf{63.3871} & 0.5511  & 34.9326 \\\hline
\textbf{Nouns}&\textbf{Good}& \textbf{270.8688} & 0.8953  & \textbf{242.5088} \\
 &\textbf{Bad}& 250.9358 & \textbf{0.9289}  & 233.0942 \\\hline
\textbf{Bigrams} &\textbf{Good}& \textbf{4116.6448} & \textbf{1.0000}  & \textbf{4116.6448} \\
 &\textbf{Bad}& 2991.6306 & \textbf{1.0000}   & 2991.6306 \\\hline
\textbf{Trigrams} &\textbf{Good}& \textbf{30424.4890} & \textbf{1.0000} & \textbf{30424.4890} \\
  &\textbf{Bad}& 17757.2356 & 0.9383  & 16661.6141 \\\hline
\textbf{Sentences} &\textbf{Good}& \textbf{8161.7926} & -0.0015 & -12.2426 \\
 &\textbf{Bad}& 2544.5235 & \textbf{0.0000}   & \textbf{0.0000}\\\hline
 \textbf{Sentences} &\textbf{Good}& 2.1199 & -0.0015 & -0.0031 \\
 \textbf{(Not Normalized)} &\textbf{Bad}& \textbf{2.1204} & \textbf{0.0000}   & \textbf{0.0000}\\\hline
\textbf{DQIC2} &\textbf{Good}& - & - & \textbf{35023.73666} \\
  &\textbf{Bad}& -& - & 20011.78371 \\\hline
  
              &            &            &        &                   
\end{tabular}
\vspace{-4mm}
\caption{Story CLOZE Sub-Component and Overall Values for $DQI_{c2}$}
\vspace{-4mm}
\end{table}

\textbf{Inter-Sample STS:}

\begin{table} [H]
\scriptsize
\centering
\begin{tabular}{llll}
\hline
\textbf{Split} & \textbf{SIML=0.3} & \textbf{SIML=0.35}& \textbf{SIML=0.4} \\
\hline
\textbf{Good} & 9.1320 & 11.3955   & 14.3267 \\
\textbf{Bad} & \textbf{10.3842} & \textbf{13.1062}   & \textbf{16.6390} \\\hline
              &            &            &                            
\end{tabular}
\vspace{-4mm}
\caption{SNLI Term 1 for $DQI_{c3}$}
\vspace{1mm}

\begin{tabular}{llll}
\hline
\textbf{Split} & \textbf{e=0.25} & \textbf{e=0.33}& \textbf{e=0.5} \\
\hline
\textbf{Good} & \textbf{0.0468} & \textbf{0.0244}   & \textbf{0.0103} \\
\textbf{Bad} & 0.0404 & 0.0216   & 0.0094 \\\hline
              &            &            &                          
\end{tabular}
\vspace{-4mm}
\caption{SNLI Term 2 for $DQI_{c3}$, with SIML=0.4}
\vspace{1mm}

\begin{tabular}{llll}
  \hline
    \multirow{2}{*}{\textbf{Sample Set}} &\multicolumn{3}{c}{\textbf{DQI C3 (e=0.5)}}\\
      &SIM=0.5&SIM=0.6&SIM=0.7\\\hline      
     \textbf{Good} &9.4123 &11.4508&14.3370  \\
     \textbf{Bad} & \textbf{10.3936} & \textbf{13.1156} & \textbf{16.7024} \\
     \hline
  \end{tabular}
  \vspace{-2mm}
 \caption{SNLI $DQI_{C3}$}
 \vspace{1mm}
\end{table}

\paragraph{}
\paragraph{}
\paragraph{}
\paragraph{}

\begin{table}
\centering
\scriptsize
\begin{tabular}{llll}
\hline
\textbf{Split} & \textbf{SIML=0.3} & \textbf{SIML=0.35}& \textbf{SIML=0.4} \\
\hline
\textbf{Good} & \textbf{334.2154} & \textbf{695.0772} & \textbf{1040.5142} \\
\textbf{Bad} & 312.4684 & 643.3308 & 953.5445 \\\hline
              &            &            &                            
\end{tabular}
\vspace{-4mm}
\caption{MNLI Term 1 for $DQI_{c3}$}
\vspace{1mm}

\begin{tabular}{llll}
\hline
\textbf{Split} & \textbf{e=0.25} & \textbf{e=0.33}& \textbf{e=0.5} \\
\hline
\textbf{Good} &  \textbf{0.0148} & \textbf{0.0108} & \textbf{0.0067} \\
\textbf{Bad} & \textbf{0.0111} & \textbf{0.0084}   & \textbf{0.0056} \\\hline
              &            &            &                          
\end{tabular}
\vspace{-4mm}
\caption{MNLI Term 2 for $DQI_{c3}$, with SIML=0.4}
\vspace{1mm}

  \begin{tabular}{llll}
  \hline
    \multirow{2}{*}{\textbf{Sample Set}} &\multicolumn{3}{c}{\textbf{DQI C3 (e=0.5)}}\\
      &SIM=0.5&SIM=0.6&SIM=0.7\\\hline      
     \textbf{Good} & \textbf{334.2221}	& \textbf{695.0839}	& \textbf{1040.5209}  \\
     \textbf{Bad} & 312.474	 & 643.3364	& 953.5501 \\
     \hline
                  &            &            &                            
  \end{tabular}
  \vspace{-4mm}
 \caption{MNLI $DQI_{C3}$} 
\vspace{1mm}

\begin{tabular}{llll}
\hline
\textbf{Split} & \textbf{SIML=0.3} & \textbf{SIML=0.35}& \textbf{SIML=0.4} \\
\hline
\textbf{Good} & \textbf{129.8631} & \textbf{171.7117} & \textbf{228.9109} \\
\textbf{Bad} & 88.9812 & 110.6097 & 141.2737 \\\hline
              &            &            &                            
\end{tabular}
  \vspace{-4mm}
\caption{SQUAD 2.0 Term 1 for $DQI_{c3}$}
  \vspace{1mm}

\begin{tabular}{llll}
\hline
\textbf{Split} & \textbf{e=0.25} & \textbf{e=0.33}& \textbf{e=0.5} \\
\hline
\textbf{Good} & 0.0051 & 0.0039   & 0.0026 \\
\textbf{Bad} & \textbf{0.0055} & \textbf{0.0042}   & \textbf{0.0094} \\\hline
              &            &            &                          
\end{tabular}
  \vspace{-4mm}
\caption{SQUAD 2.0 Term 2 for $DQI_{c3}$, with SIML=0.4}
  \vspace{1mm}

  \begin{tabular}{llll}
  \hline
    \multirow{2}{*}{\textbf{Sample Set}} &\multicolumn{3}{c}{\textbf{DQI C3 (e=0.5)}}\\
      &SIM=0.5&SIM=0.6&SIM=0.7\\\hline      
     \textbf{Good} & \textbf{129.8657} &	\textbf{171.7143} &	\textbf{228.9135}  \\
     \textbf{Bad} & 88.984 &	110.6125 &	141.2765 \\
     \hline
              &            &            &                          
  \end{tabular}
  \vspace{-4mm}
 \caption{SQUAD 2.0 $DQI_{C3}$}
\vspace{1mm}

  \begin{tabular}{llll}
\hline
\textbf{Split} & \textbf{SIML=0.3} & \textbf{SIML=0.35}& \textbf{SIML=0.4} \\
\hline
\textbf{Good} & \textbf{285.1348} & \textbf{513.1720} & \textbf{820.2516} \\
\textbf{Bad} & 209.0823 & 368.5646 & 594.0969 \\\hline
              &            &            &                            
\end{tabular}
  \vspace{-4mm}
\caption{Story CLOZE Term 1 for $DQI_{c3}$}
  \vspace{1mm}

\begin{tabular}{llll}
\hline
\textbf{Split} & \textbf{e=0.25} & \textbf{e=0.33}& \textbf{e=0.5} \\
\hline
\textbf{Good} & \textbf{0.0069} & \textbf{0.0053}   & \textbf{0.0036} \\
\textbf{Bad} & \textbf{0.0069} & \textbf{0.0053}   & \textbf{0.0036} \\\hline
              &            &            &                          
\end{tabular}
  \vspace{-4mm}
\caption{Story CLOZE Term 2 for $DQI_{c3}$, with SIML=0.4}
  \vspace{1mm}

  \begin{tabular}{llll}
  \hline
    \multirow{2}{*}{\textbf{Sample Set}} &\multicolumn{3}{c}{\textbf{DQI C3 (e=0.5)}}\\
      &SIM=0.5&SIM=0.6&SIM=0.7\\\hline      
     \textbf{Good} & \textbf{285.1384} &	\textbf{513.1756} &	\textbf{820.2552}  \\
     \textbf{Bad} & 209.0859 &	368.5682 &	594.1005 \\
     \hline
              &            &            &                          
  \end{tabular}
   \vspace{-4mm}
 \caption{Story CLOZE $DQI_{C3}$}
\vspace{-2mm}
\end{table}

\textbf{Intra-Sample Word Similarity:}
\vspace{-2mm}
\begin{table} [H]
\centering
\scriptsize
\begin{tabular}{ccc}
\hline
\textbf{Split} & \textbf{DQIC4}  \\
\hline
\textbf{Good} & \textbf{0.0004} \\
\textbf{Bad} & 0.0001 \\\hline
              &            &                 
\end{tabular}
   \vspace{-4mm}
\caption{ SNLI $DQI_{c4}$}
\label{tab:my-tablec4}
   \vspace{1mm}

\begin{tabular}{ccc}
\hline
\textbf{Split} & \textbf{DQIC4}  \\
\hline
\textbf{Good} & \textbf{0.0197} \\
\textbf{Bad} & 0.0011 \\\hline
              &            &                 
\end{tabular}
   \vspace{-4mm}
\caption{MNLI $DQI_{c4}$}
   \vspace{1mm}
 
\begin{tabular}{ccc}
\hline
\textbf{Split} & \textbf{DQIC4}  \\
\hline
\textbf{Good} & \textbf{5.2208} \\
\textbf{Bad} & 0.4577 \\\hline
              &            &                 
\end{tabular}
   \vspace{-4mm}
\caption{SQUAD 2.0 $DQI_{c4}$}
   \vspace{1mm}

\begin{tabular}{ccc}
\hline
\textbf{Split} & \textbf{DQIC4}  \\
\hline
\textbf{Good} & \textbf{0.0025} \\
\textbf{Bad} & 0.0008 \\\hline
              &            &                 
\end{tabular}
   \vspace{-4mm}
\caption{Story CLOZE $DQI_{c4}$}
\end{table}

\textbf{Intra-Sample STS:}

\begin{table} [H]
\centering
\scriptsize
\begin{tabular}{lllll}
\hline
\textbf{Split} & \textbf{ISIM=0.3}& \textbf{ISIM=0.4}& \textbf{ISIM=0.5}& \textbf{ISIM=0.6}\\
\hline
\textbf{Good} & \textbf{2.2349} & \textbf{2.8763} & \textbf{4.0125} & \textbf{6.3065}\\
\textbf{Bad}& 2.2215 & 2.8558 & 3.9784 & 6.2237 \\\hline
              &            &            &            &              
\end{tabular}
   \vspace{-4mm}
\caption{SNLI Term 1 for $DQI_{c5}$}
   \vspace{1mm}

\begin{tabular}{llllll}
\hline
\textbf{Split} & \textbf{T2}& \textbf{T3}& \textbf{T4}& \textbf{T5}& \textbf{T6} \\
\hline
\textbf{Good} & \textbf{0.1439} & \textbf{0.0038} & \textbf{6.4064e-05}& \textbf{20.3518}& \textbf{0.0903}\\
\textbf{Bad} & 0.1430& 0.0007& 1.2711e-05&19.9288 &0.0900 \\\hline
              &   & &   & &   
\end{tabular}
   \vspace{-4mm}
\caption{SNLI Terms 2,3,4,5,6 for $DQI_{c5}$}
   \vspace{1mm}
   
\begin{tabular}{ll}
\hline
\textbf{Split} & \textbf{DQI C5} \\
\hline
\textbf{Good} & \textbf{24.6024}\\
\textbf{Bad} & 24.1409\\\hline
              &                                      
\end{tabular}
   \vspace{-4mm}
\caption{SNLI $DQI_{c5}$, with ISIM=0.5}
\label{tab:my-tablec5}
   \vspace{1mm}

\begin{tabular}{lllll}
\hline
\textbf{Split} & \textbf{ISIM=0.3}& \textbf{ISIM=0.4}& \textbf{ISIM=0.5}& \textbf{ISIM=0.6}\\
\hline
\textbf{Good} & \textbf{2.2233} &  \textbf{2.8585} &   \textbf{3.9884} &   \textbf{6.3364} \\
\textbf{Bad}& 2.1256 & 2.6986 & 3.6843 & 5.5845 \\\hline
              &            &            &            &              
\end{tabular}
   \vspace{-4mm}
\caption{MNLI Term 1 for $DQI_{c5}$}
   \vspace{1mm}

\begin{tabular}{llllll}
\hline
\textbf{Split} & \textbf{T2}& \textbf{T3}& \textbf{T4}& \textbf{T5}& \textbf{T6} \\
\hline
\textbf{Good} & \textbf{0.0791} & 0.0162 & 1.1073E-05 & \textbf{15.3835}& 14.7547\\
\textbf{Bad} & 0.0741& \textbf{0.0307} & \textbf{20.9407E-05} & 12.3932 & \textbf{17.6181} \\\hline
              &   & &   & &   
\end{tabular}
   \vspace{-4mm}
\caption{MNLI Terms 2,3,4,5,6 for $DQI_{c5}$}
   \vspace{1mm}

\begin{tabular}{ll}
\hline
\textbf{Split} & \textbf{DQI C5} \\
\hline
\textbf{Good} & \textbf{34.2219}\\
\textbf{Bad} & 33.8006\\\hline
              &                                      
\end{tabular}
   \vspace{-4mm}
\caption{MNLI $DQI_{c5}$, with ISIM=0.5}
   \vspace{1mm}

\begin{tabular}{lllll}
\hline
\textbf{Split} & \textbf{ISIM=0.3}& \textbf{ISIM=0.4}& \textbf{ISIM=0.5}& \textbf{ISIM=0.6}\\
\hline
\textbf{Good} & 2.5073 & 3.3460 & 5.0031 &     9.1300\\
\textbf{Bad}&  \textbf{2.5379} & \textbf{3.4012} & \textbf{5.1352} & \textbf{9.6189}\\\hline
              &            &            &            &              
\end{tabular}
   \vspace{-4mm}
\caption{SQUAD 2.0 Term 1 for $DQI_{c5}$}
   \vspace{1mm}

\begin{tabular}{llllll}
\hline
\textbf{Split} & \textbf{T2}& \textbf{T3}& \textbf{T4}& \textbf{T5}& \textbf{T6} \\
\hline
\textbf{Good} & \textbf{0.0085} & 0.0052 & 7.3081E-06 & 22.9314 &  \textbf{102.9990}\\
\textbf{Bad} & 0.0079 & \textbf{0.0524} & \textbf{7.4403E-05} & \textbf{27.0966} & 88.8872 \\\hline
              &   & &   & &   
\end{tabular}
   \vspace{-4mm}
\caption{SQUAD 2.0 Terms 2,3,4,5,6 for $DQI_{c5}$}
   \vspace{1mm}

\begin{tabular}{ll}
\hline
\textbf{Split} & \textbf{DQI C5} \\
\hline
\textbf{Good} & \textbf{130.9472} \\
\textbf{Bad} & 121.1793\\\hline
              &                                      
\end{tabular}
   \vspace{-4mm}
\caption{SQUAD 2.0 $DQI_{c5}$, with ISIM=0.5}
   \vspace{1mm}

\begin{tabular}{lllll}
\hline
\textbf{Split} & \textbf{ISIM=0.3}& \textbf{ISIM=0.4}& \textbf{ISIM=0.5}& \textbf{ISIM=0.6}\\
\hline
\textbf{Good} &  \textbf{3.1103} & \textbf{4.5013} &  \textbf{7.7337} &   14.4898 \\
\textbf{Bad} & 3.0639 & 4.4163 &  7.5943 & \textbf{14.7772}     \\\hline
              &            &            &            &              
\end{tabular}
\vspace{-4mm}
\caption{Story CLOZE Term 1 for $DQI_{c5}$}
   \vspace{1mm}

\begin{tabular}{llllll}
\hline
\textbf{Split} & \textbf{T2}& \textbf{T3}& \textbf{T4}& \textbf{T5}& \textbf{T6} \\
\hline
\textbf{Good} &  \textbf{0.0400} & 0.0027 & 3.1939E-05  &   \textbf{0.0400} & 2.6196e-06\\
\textbf{Bad} & 0.0398 & \textbf{0.0084} & \textbf{9.7664E-05} & 0.0398 &  \textbf{7.6306e-06}  \\\hline
              &   & &   & &   
\end{tabular}
\vspace{-4mm}
\caption{Story CLOZE Terms 2,3,4,5,6 for $DQI_{c5}$}
   \vspace{1mm}

\begin{tabular}{ll}
\hline
\textbf{Split} & \textbf{DQI C5} \\
\hline
\textbf{Good} & \textbf{7.8164} \\
\textbf{Bad} & 7.6824\\\hline
              &                                      
\end{tabular}
\vspace{-4mm}
\caption{Story CLOZE $DQI_{c5}$, with ISIM=0.5}
\vspace{-4mm}
\end{table}

\textbf{N-Gram Frequency Per Label}
\begin{table} [H]
\centering
\scriptsize
\begin{tabular}{ccccc}
\hline
\textbf{Split/Label} & \textbf{Entailment} & \textbf{Neutral}& \textbf{Contradiction} \\
\hline
\textbf{Good} & 1110 & 1430   & 708 \\
\textbf{Bad} & 5626 & 5008   & 6118 \\\hline
              &            &            &       &                    
\end{tabular}
\vspace{-4mm}
\caption{SNLI Sample counts for Splits across Labels}
\label{tab:dqi6spilt}
\end{table}

\begin{table} [H]
\centering
\scriptsize
\begin{tabular}{llll}
\hline
\textbf{Split-Label} & \textbf{T1}& \textbf{T2} \\
\hline
\textbf{Good-Entailment}  & 8829.2425 & \textbf{0.9387}\\
\textbf{Bad-Entailment}  & \textbf{21655.2868} & 0.8571\\
\textbf{Good-Neutral}  & 7467.5349 & 0.8699\\
\textbf{Bad-Neutral}  & \textbf{31616.2545} & \textbf{0.9141}\\
\textbf{Good-Contradiction} & 4932.7421 & \textbf{0.9210} \\
\textbf{Bad-Contradiction} & \textbf{29145.0957} & 0.8783 \\\hline
              &            &            &                          
\end{tabular}
\vspace{-4mm}
\caption{SNLI Terms 1 and 2 for $DQI_{c6}$, Sentence Granularity}
\label{tab:dqi61sent}
\vspace{1mm}

\begin{tabular}{llll}
\hline
\textbf{Split-Label}  & \textbf{T1}& \textbf{T2} \\
\hline
\textbf{Good-Entailment}  & \textbf{142.8571} & \textbf{0.7277}\\
\textbf{Bad-Entailment}  & 81.9672 & 0.6110\\
\textbf{Good-Neutral}  & \textbf{153.8462} & \textbf{0.9118}\\
\textbf{Bad-Neutral}  & 117.6471 & 0.7071\\
\textbf{Good-Contradiction}  & \textbf{163.9344} & \textbf{0.6764} \\
\textbf{Bad-Contradiction}  & 101.0101 & 0.6088 \\\hline
              &            &            &                          
\end{tabular}
\vspace{-4mm}
\caption{SNLI Terms 1 and 2 for $DQI_{c6}$, Word Granularity}
\label{tab:dqi61word}
\vspace{1mm}

\begin{tabular}{llll}
\hline
\textbf{Split-Label}  & \textbf{T1}& \textbf{T2} \\
\hline
\textbf{Good-Entailment}  & \textbf{42.1230} & \textbf{0.34114}\\
\textbf{Bad-Entailment}  & 26.4201 & 0.30551\\
\textbf{Good-Neutral}  & \textbf{48.8998} & 0.46865\\
\textbf{Bad-Neutral}  & 38.1534 & \textbf{0.47497}\\
\textbf{Good-Contradiction}  & \textbf{43.1593} & 0.31019 \\
\textbf{Bad-Contradiction}  & 29.2826 & \textbf{0.32385} \\\hline
              &            &            &                          
\end{tabular}
\vspace{-4mm}
\caption{SNLI Terms 1 and 2 for $DQI_{c6}$, Adjective Granularity}
\label{tab:dqi61adj}
\vspace{1mm}

\begin{tabular}{llll}
\hline
\textbf{Split-Label}  & \textbf{T1}& \textbf{T2} \\
\hline
\textbf{Good-Entailment}  & \textbf{18.4128} & 0.056911\\
\textbf{Bad-Entailment}  & 11.0963 & \textbf{0.05816}\\
\textbf{Good-Neutral}  & 8.6798 & 0.09709\\
\textbf{Bad-Neutral}  & \textbf{14.6135} & \textbf{0.43124}\\
\textbf{Good-Contradiction}  & \textbf{37.9795} & \textbf{0.34286} \\
\textbf{Bad-Contradiction}  & 23.7192 & 0.21583 \\\hline
              &            &            &                          
\end{tabular}
\vspace{-4mm}
\caption{SNLI Terms 1 and 2 for $DQI_{c6}$, Adverb Granularity}
\vspace{1mm}

\begin{tabular}{llll}
\hline
\textbf{Split-Label}  & \textbf{T1}& \textbf{T2} \\
\hline
\textbf{Good-Entailment}  & \textbf{41.7885} & \textbf{0.16091}\\
\textbf{Bad-Entailment}  & 22.9410 & 0.05348\\
\textbf{Good-Neutral}  & \textbf{48.9476} & 0.17946\\
\textbf{Bad-Neutral}  & 38.9105 & \textbf{0.20192}\\
\textbf{Good-Contradiction} & \textbf{53.5045} & \textbf{0.20000} \\
\textbf{Bad-Contradiction}  & 34.6380 & 0.13589 \\\hline
              &            &            &                          
\end{tabular}
\vspace{-4mm}
\caption{SNLI Terms 1 and 2 for $DQI_{c6}$, Verb Granularity}
\vspace{1mm}

\begin{tabular}{llll}
\hline
\textbf{Split-Label}  & \textbf{T1}& \textbf{T2} \\
\hline
\textbf{Good-Entailment}  & \textbf{59.2768} & \textbf{0.49650}\\
\textbf{Bad-Entailment}  & 34.3643 & 0.38238\\
\textbf{Good-Neutral}  & \textbf{62.7353} & \textbf{0.44534}\\
\textbf{Bad-Neutral}  & 46.4253 & 0.40586\\
\textbf{Good-Contradiction}  & \textbf{66.3570} & \textbf{0.45653} \\
\textbf{Bad-Contradiction}  & 39.9202 & 0.37431 \\\hline
              &            &            &                          
\end{tabular}
\vspace{-4mm}
\caption{SNLI Terms 1 and 2 for $DQI_{c6}$, Noun Granularity}
\vspace{1mm}
\label{tab:dqi61noun}

\begin{tabular}{llll}
\hline
\textbf{Split-Label}  & \textbf{T1}& \textbf{T2} \\
\hline
\textbf{Good-Entailment}  & 1131.7133 & \textbf{0.93307}\\
\textbf{Bad-Entailment}  & \textbf{1173.5409} & 0.93206\\
\textbf{Good-Neutral}  & 1261.2663 & 0.93783\\
\textbf{Bad-Neutral}  & \textbf{1598.1514} & \textbf{0.94117}\\
\textbf{Good-Contradiction}  & 1100.8597 & \textbf{0.94325} \\
\textbf{Bad-Contradiction}  & \textbf{1369.0528} & 0.93387 \\\hline
              &            &            &                          
\end{tabular}
\vspace{-4mm}
\caption{SNLI Terms 1 and 2 for $DQI_{c6}$, Bigram Granularity}
\vspace{1mm}
\label{tab:dqi61bigram}

\begin{tabular}{llll}
\hline
\textbf{Split-Label}  & \textbf{T1}& \textbf{T2} \\
\hline
\textbf{Good-Entailment}  & 5921.2942 & \textbf{0.94672}\\
\textbf{Bad-Entailment}  & \textbf{7757.5306} & 0.93496\\
\textbf{Good-Neutral}  & 6414.8208 & 0.94517\\
\textbf{Bad-Neutral}  & \textbf{10229.7186} & \textbf{0.95015}\\
\textbf{Good-Contradiction}  & 5478.1014 & \textbf{0.95359} \\
\textbf{Bad-Contradiction}  & \textbf{8984.3224} & 0.94430 \\\hline
              &            &            &                          
\end{tabular}
\vspace{-4mm}
\caption{SNLI Terms 1 and 2 for $DQI_{c6}$, Trigram Granularity}
\vspace{-4mm}
\end{table}

\begin{table} [H]
\centering
\scriptsize
\begin{tabular}{lllllll}
\hline
\textbf{Split-Repetition} & \textbf{1} & \textbf{2}& \textbf{3} & \textbf{4}& \textbf{5}& \textbf{6}\\
\hline
\textbf{Good-Entailment} & 0.9844 & 0.0155 & 0 & 0 & 0 & 0 \\
\textbf{Bad-Entailment} &  0.9659& 0.0308&    0.001849&    0 &0.0007&    0.0005 \\
\textbf{Good-Neutral} &0.9667&    0.0325&    0.0007& 0 & 0 &0 \\
\textbf{Bad-Neutral} &0.9785&    0.0204&    0.0010& 0 & 0 & 0 \\
\textbf{Good-Contradiction} &    0.9798&    0.0201  & 0 & 0 & 0 & 0 \\
\textbf{Bad-Contradiction} &     0.9785&    0.0204&    0.0010 & 0 & 0 & 0 \\\hline
              &      &      &    &    &    &      
\end{tabular}
\vspace{-4mm}
\caption{SNLI Sentence Granularity Repetitions}
\label{tab:dqi6sentrep}
\vspace{1mm}

\begin{tabular}{ll}
\hline
\textbf{Split-Label} & \textbf{T3}\\
\hline
\textbf{Good-Entailment} & \textbf{0.1457} \\
\textbf{Bad-Entailment} & 0.1330 \\
\textbf{Good-Neutral} & 0.1496 \\
\textbf{Bad-Neutral} & \textbf{0.1571} \\
\textbf{Good-Contradiction} & 0.1313 \\
\textbf{Bad-Contradiction} & \textbf{0.1434} \\\hline
              &                                    
\end{tabular}
\vspace{-4mm}
\caption{SNLI T3 for $DQI_{c6}$}
\label{tab:dqi63}
\vspace{1mm}

\begin{tabular}{ll}
\hline
\textbf{Split-Label} & \textbf{T4}\\
\hline
\textbf{Good-Entailment} & \textbf{0.0100} \\
\textbf{Bad-Entailment} & 0.0021 \\
\textbf{Good-Neutral} & \textbf{0.0084} \\
\textbf{Bad-Neutral} & 0.0022 \\
\textbf{Good-Contradiction} & 0.0197 \\
\textbf{Bad-Contradiction} & \textbf{0.0020} \\\hline
              &                                    
\end{tabular}
\vspace{-4mm}
\caption{SNLI T4 for $DQI_{c6}$}
\label{tab:dqi64}
\vspace{1mm}

\begin{tabular}{lll}
\hline
\textbf{Granularity/Split} & \textbf{Good}& \textbf{Bad}\\
\hline
\textbf{Sentences} & \textbf{15.3475} & 11.6614\\
\textbf{Words} & \textbf{0.9313} & 0.6596  \\
\textbf{Adjectives} & \textbf{1.2190} & 0.9185  \\
\textbf{Adverbs} & \textbf{1.5708} & 1.1850 \\
\textbf{Verbs} & \textbf{0.9667} & 0.7001 \\
\textbf{Nouns} & \textbf{1.0623} & 0.7358  \\
\textbf{Bigrams} & 0.3646 & \textbf{0.4893} \\
\textbf{Trigrams} & 0.1860 & \textbf{0.2760} \\\hline
              &      &
\end{tabular}
\vspace{-4mm}
\caption{SNLI T5 for $DQI_{c6}$}
\label{tab:dqi65}
\vspace{1mm}

\begin{tabular}{ll}
\hline
\textbf{Split-Label} & \textbf{DQI C6}\\
\hline
\textbf{Good} & \textbf{556.6914} \\
\textbf{Bad} &  320.2893\\\hline
              &                                    
\end{tabular}
\vspace{-4mm}
\caption{SNLI $DQI_{c6}$}
\label{tab:dqic6final}
\vspace{1mm}

\begin{tabular}{ccccc}
\hline
\textbf{Split/Label} & \textbf{Entailment} & \textbf{Neutral}& \textbf{Contradiction} \\
\hline
\textbf{Good} & 6150 & 6098 & 6082 \\
\textbf{Bad} & 700 & 60 & 240 \\\hline
              &            &            &       &                    
\end{tabular}
\vspace{-4mm}
\caption{MNLI Sample counts for Splits across Labels}
\vspace{1mm}

\begin{tabular}{llll}
\hline
\textbf{Split-Label} & \textbf{T1}& \textbf{T2} \\
\hline
\textbf{Good-Entailment}  & 2.69E+04 &	0.8133\\
\textbf{Bad-Entailment}  &6.47E+03 &	0.9542\\
\textbf{Good-Neutral}  &2.78E+04 &	0.8465\\
\textbf{Bad-Neutral}  &4.76E+16 &	1.0000\\
\textbf{Good-Contradiction}  &4.62E+04 &	0.9378\\
\textbf{Bad-Contradiction}  &1.05E+17 &	1.0000\\\hline
              &            &            &                          
\end{tabular}
\vspace{-4mm}
\caption{MNLI Terms 1 and 2 for $DQI_{c6}$, Sentence Granularity}
\vspace{1mm}

\begin{tabular}{llll}
\hline
\textbf{Split-Label}  & \textbf{T1}& \textbf{T2} \\
\hline
\textbf{Good-Entailment}&5.67E+02&	0.970607701\\
\textbf{Bad-Entailment}&9.48E+02&	0.957116548\\
\textbf{Good-Neutral}&8.70E+02&	0.488048002\\
\textbf{Bad-Neutral}&6.74E+02&	0.794573643\\
\textbf{Good-Contradiction}&9.40E+02&	0.965482191\\
\textbf{Bad-Contradiction}&7.01E+02&	0.885763001\\\hline
              &            &            &                          
\end{tabular}
\vspace{-4mm}
\caption{MNLI Terms 1 and 2 for $DQI_{c6}$, Word Granularity}
\vspace{1mm}

\begin{tabular}{llll}
\hline
\textbf{Split-Label}  & \textbf{T1}& \textbf{T2} \\
\hline
\textbf{Good-Entailment}&1.16E+02&	0.7834\\
\textbf{Bad-Entailment}&2.83E+02&	1.0000\\
\textbf{Good-Neutral}&2.86E+02&	1.0000\\
\textbf{Bad-Neutral}&1.92E+02&	0.8771\\
\textbf{Good-Contradiction}&3.47E+02&	1.0000\\
\textbf{Bad-Contradiction}&2.67E+02&	1.0000\\\hline
              &            &            &                          
\end{tabular}
\vspace{-4mm}
\caption{MNLI Terms 1 and 2 for $DQI_{c6}$, Adjective Granularity}
\vspace{-4mm}
\end{table}

\begin{table} [H]
\centering
\scriptsize
\begin{tabular}{llll}
\hline
\textbf{Split-Label}  & \textbf{T1}& \textbf{T2} \\
\hline
\textbf{Good-Entailment}&2.56E+01&	0.4803\\
\textbf{Bad-Entailment}&5.20E+01&	0.6531\\
\textbf{Good-Neutral}&3.61E+01&	0.6091\\
\textbf{Bad-Neutral}&7.15E+01&	0.6521\\
\textbf{Good-Contradiction}&3.43E+01&	0.5017\\
\textbf{Bad-Contradiction}&5.19E+01&	0.3939\\\hline
              &            &            &                          
\end{tabular}
\vspace{-4mm}
\caption{MNLI Terms 1 and 2 for $DQI_{c6}$, Adverb Granularity}
\vspace{1mm}

\begin{tabular}{llll}
\hline
\textbf{Split-Label}  & \textbf{T1}& \textbf{T2} \\
\hline
\textbf{Good-Entailment}&1.71E+02&	0.7901\\
\textbf{Bad-Entailment}&1.61E+02&	0.6620\\
\textbf{Good-Neutral}&1.43E+02&	0.5911\\
\textbf{Bad-Neutral}&1.69E+02&	0.3061\\
\textbf{Good-Contradiction}&1.79E+02&	0.7271\\
\textbf{Bad-Contradiction}&1.30E+02&	0.5636\\\hline
              &            &            &                          
\end{tabular}
\vspace{-4mm}
\caption{MNLI Terms 1 and 2 for $DQI_{c6}$, Verb Granularity}

\begin{tabular}{llll}
\hline
\textbf{Split-Label}  & \textbf{T1}& \textbf{T2} \\
\hline
\textbf{Good-Entailment}&2.61E+02&	0.8994\\
\textbf{Bad-Entailment}&4.52E+02&	0.9447\\
\textbf{Good-Neutral}&4.68E+02&	1.0000\\
\textbf{Bad-Neutral}&2.61E+02&	0.7235\\
\textbf{Good-Contradiction}&4.84E+02&	1.0000\\
\textbf{Bad-Contradiction}&2.80E+02&	0.9287\\\hline
              &            &            &                          
\end{tabular}
\vspace{-4mm}
\caption{MNLI Terms 1 and 2 for $DQI_{c6}$, Noun Granularity}
\vspace{1mm}

\begin{tabular}{llll}
\hline
\textbf{Split-Label}  & \textbf{T1}& \textbf{T2} \\
\hline
\textbf{Good-Entailment}&3.38E+03&	0.9361\\
\textbf{Bad-Entailment}&4.83E+03&	1.0000\\
\textbf{Good-Neutral}&9.21E+03&	1.0000\\
\textbf{Bad-Neutral}&1.91E+03&	1.0000\\
\textbf{Good-Contradiction}&1.04E+04&	1.0000\\
\textbf{Bad-Contradiction}&2.97E+03&	1.0000\\\hline

              &            &            &                          
\end{tabular}
\vspace{-4mm}
\caption{MNLI Terms 1 and 2 for $DQI_{c6}$, Bigram Granularity}
\vspace{1mm}

\begin{tabular}{llll}
\hline
\textbf{Split-Label}  & \textbf{T1}& \textbf{T2} \\
\hline
\textbf{Good-Entailment}&9.27E+03&	0.9573\\
\textbf{Bad-Entailment}&2.93E+04&	1.0000\\
\textbf{Good-Neutral}&4.54E+04&	0.9913\\
\textbf{Bad-Neutral}&4.61E+03&	0.8822\\
\textbf{Good-Contradiction}&1.04E+05&	1.0000\\
\textbf{Bad-Contradiction}&6.96E+03&	0.9937\\\hline
              &            &            &                          
\end{tabular}
\vspace{-4mm}
\caption{MNLI Terms 1 and 2 for $DQI_{c6}$, Trigram Granularity}
\vspace{1mm}

\begin{tabular}{llll}
\hline
\textbf{Split-Repetition} & \textbf{1} & \textbf{2}& \textbf{3} \\
\hline
\textbf{Good-Entailment} & 0.9512 & 0.0484 & 0.0003  \\
\textbf{Bad-Entailment} &  0.9884 & 0.0115 &    0.0000  \\
\textbf{Good-Neutral} & 0.9612 &    0.0363 &    0.0024  \\
\textbf{Bad-Neutral} & 1.0000 &    0.0000 &    0.0000  \\
\textbf{Good-Contradiction} &    0.9844 &    0.0150  & 0.0005  \\
\textbf{Bad-Contradiction} &     1.0000 &    0.0000 & 0.0000  \\\hline
              &      &      &    
\end{tabular}
\vspace{-4mm}
\caption{MNLI Sentence Granularity Repetitions}
\vspace{1mm}

\begin{tabular}{ll}
\hline
\textbf{Split-Label} & \textbf{T3}\\
\hline
\textbf{Good-Entailment}&0.0647\\
\textbf{Bad-Entailment}&0.0860\\
\textbf{Good-Neutral}&0.0926\\
\textbf{Bad-Neutral}&0.0590\\
\textbf{Good-Contradiction}&0.1000\\
\textbf{Bad-Contradiction}&0.2290\\\hline
              &                                    
\end{tabular}
\vspace{-4mm}
\caption{MNLI T3 for $DQI_{c6}$}
\vspace{1mm}

\begin{tabular}{ll}
\hline
\textbf{Split-Label} & \textbf{T4}\\
\hline
\textbf{Good-Entailment}&0.0803\\
\textbf{Bad-Entailment}&0.0202\\
\textbf{Good-Neutral}&0.0041\\
\textbf{Bad-Neutral}&0.0484\\
\textbf{Good-Contradiction}&0.2018\\
\textbf{Bad-Contradiction}&0.0326\\\hline
              &                                    
\end{tabular}
\vspace{-4mm}
\caption{MNLI T4 for $DQI_{c6}$}
\vspace{1mm}

\begin{tabular}{ll}
\hline
\textbf{Split-Label} & \textbf{DQI C6}\\
\hline
\textbf{Good} & 2.74E+05 \\
\textbf{Bad} &  1.53E+17\\\hline
              &                                    
\end{tabular}
\vspace{-4mm}
\caption{MNLI $DQI_{c6}$}
\vspace{-4mm}
\end{table}

\begin{table} [H]
\centering
\scriptsize
\begin{tabular}{lll}
\hline
\textbf{Granularity/Split} & \textbf{Good}& \textbf{Bad}\\
\hline
\textbf{Sentences} & 14.6049 & 72.1687\\
\textbf{Words} & 1.2571 & 0.8533  \\
\textbf{Adjectives} & 1.4557 &	1.7959  \\
\textbf{Adverbs} & 0.7319	& 0.9429 \\
\textbf{Verbs} & 1.0382	& 1.0345 \\
\textbf{Nouns} & 1.7755	& 1.5836  \\
\textbf{Bigrams} & 0.4008	& 0.3561 \\
\textbf{Trigrams} & 0.6547	& 0.9724 \\\hline
              &      &
\end{tabular}
\vspace{-4mm}
\caption{MNLI T5 for $DQI_{c6}$}
\vspace{1mm}

\begin{tabular}{ccc}
\hline
\textbf{Split/Label} & \textbf{True} & \textbf{False} \\
\hline
\textbf{Good} & 10946 & 10770 \\
\textbf{Bad} & 914 & 1086 \\\hline
              &            &                             
\end{tabular}
\vspace{-4mm}
\caption{SQUAD 2.0 Sample counts for Splits across Labels}
\vspace{1mm}

\begin{tabular}{llll}
\hline
\textbf{Split-Label} & \textbf{T1}& \textbf{T2} \\
\hline
\textbf{Good-True} &4431.2159 & 0.0007\\
\textbf{Bad-True}&1921.2260  &0.5448 \\
\textbf{Good-False}&4412.2037  &0.0014 \\
\textbf{Bad-False}&1853.6963  &0.5009\\\hline
              &            &            &                          
\end{tabular}
\vspace{-4mm}
\caption{SQUAD 2.0 Terms 1 and 2 for $DQI_{c6}$, Sentence Granularity}
\vspace{1mm}

\begin{tabular}{llll}
\hline
\textbf{Split-Label}  & \textbf{T1}& \textbf{T2} \\
\hline
\textbf{Good-True} &263.6776&	1.0000\\
\textbf{Bad-True}&954.5225&	1.0000 \\
\textbf{Good-False}&259.3381&	0.3105 \\
\textbf{Bad-False}&776.2031&	1.0000\\\hline
              &            &            &                          
\end{tabular}
\vspace{-4mm}
\caption{SQUAD 2.0 Terms 1 and 2 for $DQI_{c6}$, Word Granularity}
\vspace{1mm}

\begin{tabular}{llll}
\hline
\textbf{Split-Label}  & \textbf{T1}& \textbf{T2} \\
\hline
\textbf{Good-True} &75.3820&	1.0000\\
\textbf{Bad-True}&244.8719&	1.0000 \\
\textbf{Good-False}&70.8210&	1.0000\\
\textbf{Bad-False}&222.5754&	1.0000\\\hline
              &            &            &                          
\end{tabular}
\vspace{-4mm}
\caption{SQUAD 2.0 Terms 1 and 2 for $DQI_{c6}$, Adjective Granularity}
\vspace{1mm}

\begin{tabular}{llll}
\hline
\textbf{Split-Label}  & \textbf{T1}& \textbf{T2} \\
\hline
\textbf{Good-True}&6.31677&	0.6666\\
\textbf{Bad-True}&27.6740&	0.6494\\
\textbf{Good-False}&6.4805&	0.6632\\
\textbf{Bad-False}&24.6482&	0.6878\\\hline
              &            &            &                          
\end{tabular}
\vspace{-4mm}
\caption{SQUAD 2.0 Terms 1 and 2 for $DQI_{c6}$, Adverb Granularity}
\vspace{1mm}

\begin{tabular}{llll}
\hline
\textbf{Split-Label}  & \textbf{T1}& \textbf{T2} \\
\hline
\textbf{Good-True}&58.2850&	0.8789\\
\textbf{Bad-True}&219.8726&	0.8851\\
\textbf{Good-False}&59.0344&	0.9066\\
\textbf{Bad-False}&208.3846&	0.9113\\\hline
              &            &            &                          
\end{tabular}
\vspace{-4mm}
\caption{SQUAD 2.0 Terms 1 and 2 for $DQI_{c6}$, Verb Granularity}
\vspace{1mm}

\begin{tabular}{llll}
\hline
\textbf{Split-Label}  & \textbf{T1}& \textbf{T2} \\
\hline
\textbf{Good-True}&110.8118&	1.0000\\
\textbf{Bad-True}&415.9473&	1.0000\\
\textbf{Good-False}&109.7139&	1.0000\\
\textbf{Bad-False}&307.1137&	1.0000\\\hline
              &            &            &                          
\end{tabular}
\vspace{-4mm}
\caption{SQUAD 2.0 Terms 1 and 2 for $DQI_{c6}$, Noun Granularity}
\vspace{1mm}

\begin{tabular}{llll}
\hline
\textbf{Split-Label}  & \textbf{T1}& \textbf{T2} \\
\hline
\textbf{Good-True}&2923.9305&	0.9768\\
\textbf{Bad-True}&5800.9793&	0.9762\\
\textbf{Good-False}&2834.7978&	0.9758\\
\textbf{Bad-False}&5157.4516&	0.9749\\\hline
              &            &            &                          
\end{tabular}
\vspace{-4mm}
\caption{SQUAD 2.0 Terms 1 and 2 for $DQI_{c6}$, Bigram Granularity}
\vspace{1mm}

\begin{tabular}{llll}
\hline
\textbf{Split-Label}  & \textbf{T1}& \textbf{T2} \\
\hline
\textbf{Good-True}&35363.3144&	1.0000\\
\textbf{Bad-True}&49074.7258&	1.0000\\
\textbf{Good-False}&34076.1381&	1.0000\\
\textbf{Bad-False}&40854.1931&	1.0000\\\hline
              &            &            &                          
\end{tabular}
\vspace{-4mm}
\caption{SQUAD 2.0 Terms 1 and 2 for $DQI_{c6}$, Trigram Granularity}
\vspace{-4mm}
\end{table}

\begin{table} [H]
\centering
\scriptsize

\begin{tabular}{ll}
\hline
\textbf{Split-Label} & \textbf{T3}\\
\hline
\textbf{Good-True}&0.0085\\
\textbf{Bad-True}&0.00852\\
\textbf{Good-False}&0.0079\\
\textbf{Bad-False}&0.0078\\\hline
              &                                    
\end{tabular}
\vspace{-4mm}
\caption{SQUAD 2.0 T3 for $DQI_{c6}$}
\vspace{1mm}

\begin{tabular}{ll}
\hline
\textbf{Split-Label} & \textbf{T4}\\
\hline
\textbf{Good-True}&0.0104\\
\textbf{Bad-True}&0.0106\\
\textbf{Good-False}&0.1165\\
\textbf{Bad-False}&0.0954\\\hline
              &                                    
\end{tabular}
\vspace{-4mm}
\caption{SQUAD 2.0 T4 for $DQI_{c6}$}
\vspace{1mm}

\begin{tabular}{lll}
\hline
\textbf{Granularity/Split} & \textbf{Good}& \textbf{Bad}\\
\hline
\textbf{Sentences} & 20.5287 & 9.6533\\
\textbf{Words} & 0.0711 & 0.0682 \\
\textbf{Adjectives} &0.6497	&1.1487\\
\textbf{Adverbs} &0.4012	&0.6832 \\
\textbf{Verbs} &0.4918	&0.8153 \\
\textbf{Nouns} &0.5183	&0.9957  \\
\textbf{Bigrams} &0.1262	&0.05600 \\
\textbf{Trigrams} &0.1366	&0.09422 \\\hline
              &      &
\end{tabular}
\vspace{-4mm}
\caption{SQUAD 2.0 T5 for $DQI_{c6}$}
\vspace{1mm}

\begin{tabular}{ll}
\hline
\textbf{Split-Label} & \textbf{DQI C6}\\
\hline
\textbf{Good} & 75918.2760 \\
\textbf{Bad} &  105949.3404\\\hline
              &                                    
\end{tabular}
\vspace{-4mm}
\caption{SQUAD 2.0 $DQI_{c6}$}
\vspace{1mm}

\begin{tabular}{ccc}
\hline
\textbf{Split/Label} & \textbf{True} & \textbf{False} \\
\hline
\textbf{Good} & 2568 & 2568 \\
\textbf{Bad} & 800 & 800 \\\hline
              &            &                             
\end{tabular}
\vspace{-4mm}
\caption{Story CLOZE Sample counts for Splits across Labels}
\vspace{1mm}

\begin{tabular}{llll}
\hline
\textbf{Split-Label} & \textbf{T1}& \textbf{T2} \\
\hline
\textbf{Good-True} &1.30E+05&	0.9984\\
\textbf{Bad-True}&5.06E+16&	1.0000 \\
\textbf{Good-False}&1.30E+05&	0.9984 \\
\textbf{Bad-False}&5.06E+16&	1.0000\\\hline
              &            &            &                          
\end{tabular}
\vspace{-4mm}
\caption{Story CLOZE Terms 1 and 2 for $DQI_{c6}$, Sentence Granularity}
\vspace{1mm}

\begin{tabular}{llll}
\hline
\textbf{Split-Label}  & \textbf{T1}& \textbf{T2} \\
\hline
\textbf{Good-True} &5.47E+05&	0.9792\\
\textbf{Bad-True}&5.22E+05&	0.8618 \\
\textbf{Good-False}&5.47E+05&	0.5316 \\
\textbf{Bad-False}&4.96E+05&	0.8537\\\hline
              &            &            &                          
\end{tabular}
\vspace{-4mm}
\caption{Story CLOZE Terms 1 and 2 for $DQI_{c6}$, Word Granularity}
\vspace{1mm}

\begin{tabular}{llll}
\hline
\textbf{Split-Label}  & \textbf{T1}& \textbf{T2} \\
\hline
\textbf{Good-True} &129.1883	&0.7800\\
\textbf{Bad-True}&133.5904&	0.7711 \\
\textbf{Good-False}&121.0435&	0.7459\\
\textbf{Bad-False}&128.3632&	0.8014\\\hline
              &            &            &                          
\end{tabular}
\vspace{-4mm}
\caption{Story CLOZE Terms 1 and 2 for $DQI_{c6}$, Adjective Granularity}
\vspace{1mm}

\begin{tabular}{llll}
\hline
\textbf{Split-Label}  & \textbf{T1}& \textbf{T2} \\
\hline
\textbf{Good-True}&41.1600	    &0.5959\\
\textbf{Bad-True}&49.9482	 &   0.5368\\
\textbf{Good-False}&36.9653	 &   0.6145\\
\textbf{Bad-False}&54.7544	 &   0.6194\\\hline
              &            &            &                          
\end{tabular}
\vspace{-4mm}
\caption{Story CLOZE Terms 1 and 2 for $DQI_{c6}$, Adverb Granularity}
\vspace{1mm}

\begin{tabular}{llll}
\hline
\textbf{Split-Label}  & \textbf{T1}& \textbf{T2} \\
\hline
\textbf{Good-True}&103.8261	&0.5285\\
\textbf{Bad-True}&115.6968&	0.5828\\
\textbf{Good-False}&112.3307&	0.5946\\
\textbf{Bad-False}&113.4481&	0.5155\\\hline
              &            &            &                          
\end{tabular}
\vspace{-4mm}
\caption{Story CLOZE Terms 1 and 2 for $DQI_{c6}$, Verb Granularity}
\vspace{1mm}

\end{table}

\begin{table} [H]
\centering
\scriptsize

\begin{tabular}{llll}
\hline
\textbf{Split-Label}  & \textbf{T1}& \textbf{T2} \\
\hline
\textbf{Good-True}&551.3272&	0.8898\\
\textbf{Bad-True}&458.9138&	0.8862\\
\textbf{Good-False}&520.3204&	0.9047\\
\textbf{Bad-False}&462.2876&	0.9252\\\hline
              &            &            &                          
\end{tabular}
\vspace{-4mm}
\caption{Story CLOZE Terms 1 and 2 for $DQI_{c6}$, Noun Granularity}
\vspace{1mm}

\begin{tabular}{llll}
\hline
\textbf{Split-Label}  & \textbf{T1}& \textbf{T2} \\
\hline
\textbf{Good-True}7139.05776&	1.0000\\
\textbf{Bad-True}5158.2473&	1.0000\\
\textbf{Good-False}6941.1989&	1.0000\\
\textbf{Bad-False}5006.1656&	1.0000\\\hline
              &            &            &                          
\end{tabular}
\vspace{-4mm}
\caption{Story CLOZE Terms 1 and 2 for $DQI_{c6}$, Bigram Granularity}
\vspace{1mm}

\begin{tabular}{llll}
\hline
\textbf{Split-Label}  & \textbf{T1}& \textbf{T2} \\
\hline
\textbf{Good-True}54497.5504&	1.0000\\
\textbf{Bad-True}33876.9502&	1.0000\\
\textbf{Good-False}50906.0915&	1.0000\\
\textbf{Bad-False}33618.6103&	1.0000\\\hline
              &            &            &                          
\end{tabular}
\vspace{-4mm}
\caption{Story CLOZE Terms 1 and 2 for $DQI_{c6}$, Trigram Granularity}
\vspace{1mm}

\begin{tabular}{ll}
\hline
\textbf{Split-Label} & \textbf{T3}\\
\hline
\textbf{Good-True}&0.0085\\
\textbf{Bad-True}&0.0079\\
\textbf{Good-False}&0.0085\\
\textbf{Bad-False}&0.0078\\\hline
              &                                    
\end{tabular}
\vspace{-4mm}
\caption{Story CLOZE 2.0 T3 for $DQI_{c6}$}
\vspace{1mm}

\begin{tabular}{ll}
\hline
\textbf{Split-Label} & \textbf{T4}\\
\hline
\textbf{Good-True}&0.0104\\
\textbf{Bad-True}&0.1165\\
\textbf{Good-False}&0.0106\\
\textbf{Bad-False}&0.0954\\\hline
              &                                    
\end{tabular}
\vspace{-4mm}
\caption{Story CLOZE 2.0 T4 for $DQI_{c6}$}
\vspace{1mm}

\begin{tabular}{lll}
\hline
\textbf{Granularity/Split} & \textbf{Good}& \textbf{Bad}\\
\hline
\textbf{Sentences} & 382.2842 & 2262.7417\\
\textbf{Words} & 1.0447 & 1.0192 \\
\textbf{Adjectives} &3.9910	&5.0527 \\
\textbf{Adverbs} &1.7714	&3.1284 \\
\textbf{Verbs} &2.2377	&3.5188 \\
\textbf{Nouns} &5.8841	&7.3696 \\
\textbf{Bigrams} &1.6522 &	1.9489\\
\textbf{Trigrams} & 4.9660	& 6.8154 \\\hline
              &      &
\end{tabular}
\vspace{-4mm}
\caption{Story CLOZE T5 for $DQI_{c6}$}
\vspace{1mm}
\end{table}

\begin{table}
    \centering
\scriptsize

\begin{tabular}{ll}
\hline
\textbf{Split-Label} & \textbf{DQI C6}\\
\hline
\textbf{Good} & 1.01E+17 \\
\textbf{Bad} &  1.01E+17\\\hline
              &                                    
\end{tabular}
\vspace{-4mm}
\caption{Story CLOZE $DQI_{c6}$}
\vspace{-4mm}
\end{table}

\textbf{Inter-Split STS:}

\begin{table} [H]
\centering
\scriptsize
\begin{tabular}{llll}
\hline
\textbf{Split} & \textbf{SSMIL=0.2}& \textbf{SSMIL=0.3}& \textbf{SSMIL=0.4}\\
\hline
\textbf{Good} & \textbf{0.0031} & \textbf{0.0042} & \textbf{0.0063} \\
\textbf{Bad} & 0.0029 & 0.0040 & 0.0057 \\\hline
              &        &        &                            
\end{tabular}
\vspace{-4mm}
\caption{SNLI $DQI_{c7}$}
\vspace{1mm}

\begin{tabular}{llll}
\hline
\textbf{Split} & \textbf{SSMIL=0.2}& \textbf{SSMIL=0.3}& \textbf{SSMIL=0.4}\\
\hline
\textbf{Good} & 0.0004 & 0.0005 & 0.0002 \\
\textbf{Bad} & \textbf{0.0009} & \textbf{0.0011} & \textbf{0.0005} \\\hline
              &        &        &                            
\end{tabular}
\vspace{-4mm}
\caption{MNLI $DQI_{c7}$}
\vspace{1mm}

\end{table}

\begin{table*}
\centering
\includegraphics[width=1.8\columnwidth]{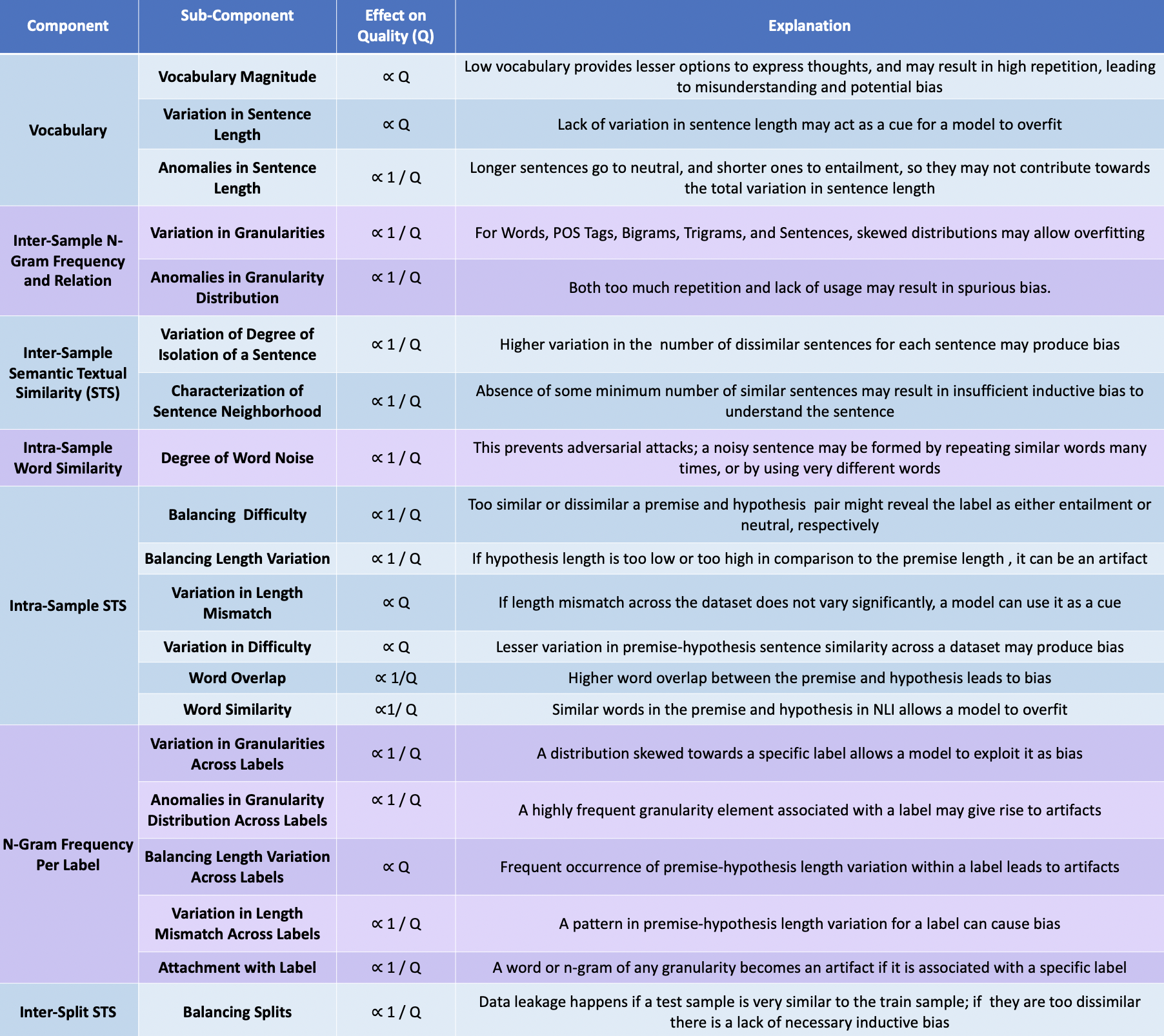}
  
  \caption{Intuitions behind DQI components and sub-components.}
\vspace{-4mm}
\end{table*}

\begin{figure*}
    \centering
    \includegraphics[width=2\columnwidth,height=6cm]{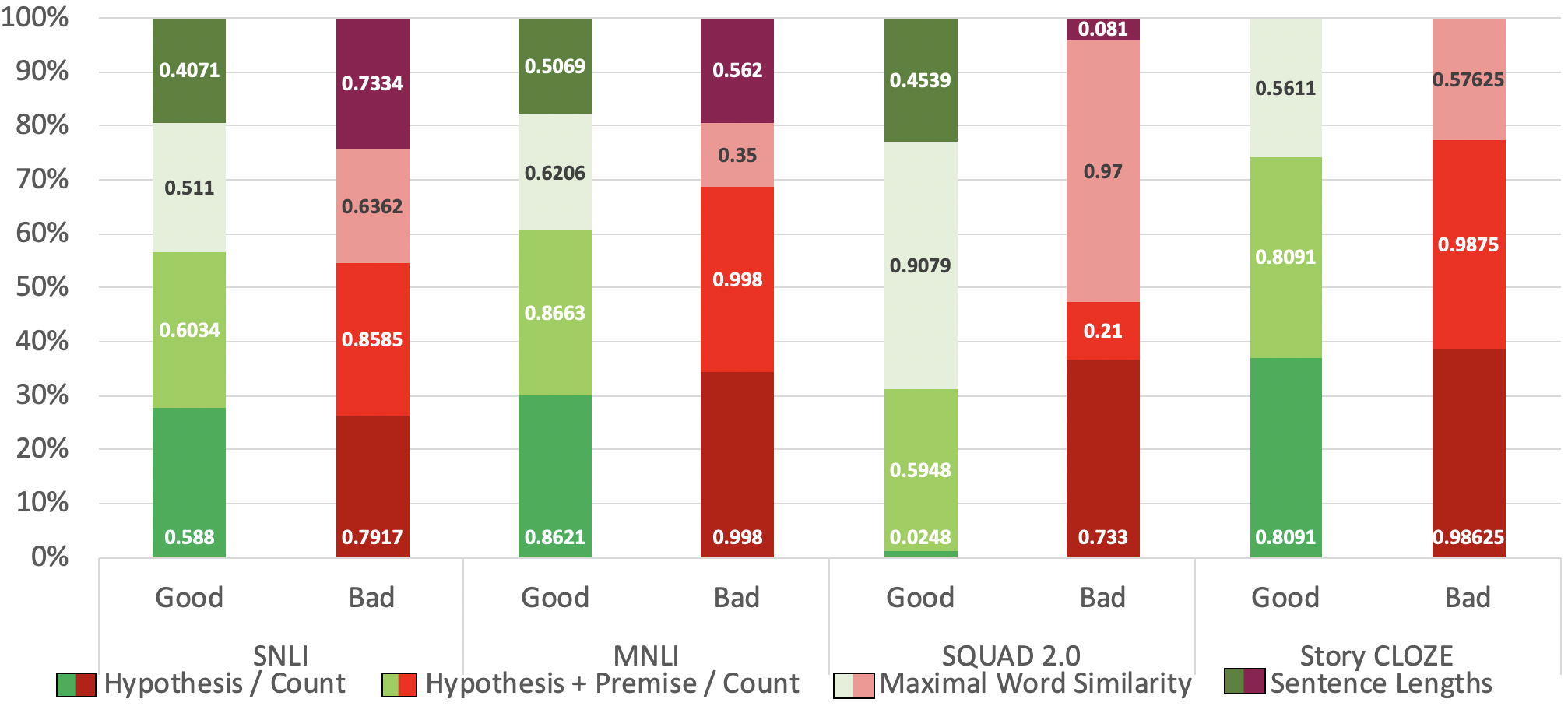}
    \centering
    \caption{
    Each bar shows the relative contribution amounts of four features: \textit{word overlap} (hypothesis only, and hypothesis$+$premise), \textit{maximal word similarity}, and \textit{sentence lengths}, for \textit{good} and \textit{bad} split samples. Each bar stacks the four features, which are sized by their relative impact percent (raw contribution values are numbers inside each feature bar).}
    \label{fig:sample}
    \vspace{-4mm}
\end{figure*}

\end{document}